\documentclass[a4paper,fleqn]{cas-dc}

\usepackage[numbers]{natbib}
\usepackage{amsmath,amssymb,amsfonts}
\usepackage{algorithmic}
\usepackage{graphicx}
\usepackage{textcomp}
\usepackage{wrapfig}
\usepackage{microtype}
\usepackage{algorithm}
\usepackage{array}
\usepackage[caption=false]{subfig}
\usepackage{url}
\usepackage{verbatim}
\usepackage{booktabs}
\usepackage{gensymb}
\usepackage[dvipsnames]{xcolor}
\usepackage{colortbl}
\usepackage{ifthen}
\usepackage{tikz}
\usepackage{float}

\usetikzlibrary{positioning, chains, shapes.geometric, fit, backgrounds, shapes, arrows.meta, calc, decorations.pathreplacing, ext.paths.ortho, quotes}
\usetikzlibrary{matrix,calc,3d}

\def\tsc#1{\csdef{#1}{\textsc{\lowercase{#1}}\xspace}}
\tsc{WGM}
\tsc{QE}

\begin{document}
\let\WriteBookmarks\relax
\def\floatpagepagefraction{1}
\def\textpagefraction{.001}

\shorttitle{}    

\shortauthors{E. Motta et~al.}  

\title [mode = title]{GenGait: A Transformer-Based Model for Human Gait Anomaly Detection and Normative Twin Generation}  

\tnotemark[1] 

\tnotetext[1]{This work was supported in part by the Italian National Institute for Insurance against Accidents at Work (INAIL) ergoCub Core Project, by the Italian Ministry of University and Research (MUR) under the Fondo Italiano per la Scienza (FIS), call FIS 3, project EPIC with code FIS-2024-02654, and by the IIT Technologies for Healthy Living Flagship.} 

%

\author[1]{Elisa Motta}
\cormark[1]


\author[1]{Marta Lorenzini}
\author[1]{Clara Mouawad}




\affiliation[1]{organization={HRI$^{2}$ Laboratory, Istituto Italiano di Tecnologia (IIT)},
            city={Genoa},
            country={Italy}}

\author[2]{Alberto Ranavolo}





\affiliation[2]{organization={Department of Occupational and Environmental Medicine, Epidemiology and Hygiene, INAIL},
            city={Rome},
            country={Italy}}

\author[3]{Mariano Serrao}

\affiliation[3]{organization={ Department of Medical and Surgical Sciences and Biotechnologies, Sapienza University of Rome},
            city={Rome},
            country={Italy}}

\author[1]{Arash Ajoudani}

\cortext[1]{Corresponding author}



\begin{abstract}
Gait analysis provides an objective characterization of locomotor function and is widely used to support diagnosis and rehabilitation monitoring across neurological and orthopedic disorders. Deep learning
has been increasingly applied to this domain, yet most approaches rely on supervised classifiers trained on disease-labeled data, limiting generalization to heterogeneous pathological presentations. The methodological objective of this work is to develop a label-free framework for joint-level anomaly detection and kinematic correction based on a Transformer masked autoencoder trained exclusively on normative gait sequences from 150 adults, acquired with a markerless multi-camera motion-capture system.

At inference, a two-pass procedure is applied to potentially pathological input sequences: first, it estimates joint inconsistency scores by occluding individual joints and measuring deviations from the learned normative prior. Then, it withholds the flagged joints from the encoder input and reconstructs the full skeleton from the remaining spatiotemporal context, yielding corrected kinematic trajectories at the flagged positions.

The validation objective is to assess whether the framework preserves unseen normative gait and reduces angular deviation in simulated abnormal gait patterns.

In this proof-of-concept evaluation, data from 10 held-out normative participants, who performed seven simulated abnormal gait patterns, showed a significant reduction in angular deviation across all analyzed joints with large effect sizes, and preservation of normative kinematics.

The proposed approach enables interpretable, subject-specific localization of joints that are inconsistent with learned normative gait patterns and generation of an individualized normative reconstruction without requiring disease labels. Video is available at \href{https://youtu.be/Rcm3jqR5pN4}{\texttt{https://youtu.be/GenGait}}.
\end{abstract}



\begin{keywords}
 Gait Analysis \sep Human Motion Analysis \sep Motion Prediction \sep Anomaly Detection \sep Transformers
\end{keywords}

\maketitle

\section{Introduction}
\label{sec:introduction}
Gait analysis is the quantitative study of human locomotion, systematically measuring kinematics, kinetics, and spatiotemporal parameters across multiple gait cycles to characterize movement quality and detect pathology \cite{whittle2014gait}. Perry and  Burnfield \cite{perry2024gait} established that normative gait is not a single standard pattern, but rather it spans a band influenced by anthropometry, age, conditioning, and individual motor strategies \cite{perry2024gait}.
Gait analysis has become the base clinical tool for understanding the locomotor function, diagnosing pathological conditions, and evaluating treatment efficacy across a wide spectrum of neurological and orthopedic disorders \cite{wren2011efficacy}.
By providing objective, reproducible measurements, it complements qualitative observational assessments, which depend on clinical expertise and are limited in sensitivity to subtle changes. 

Currently, marker-based optoelectronic motion-capture systems are the gold standard for gait analysis, enabling estimation of joint kinematics and spatiotemporal parameters \cite{kanko2021concurrent}.
When combined with force plates to measure ground reaction forces and electromyography for muscle activation patterns, these systems provide the most comprehensive kinematic and kinetic characterization currently available in clinical practice. However, despite their accuracy, laboratory workflows depend on expensive and cumbersome setups and highly standardized protocols, which can limit accessibility and ecological validity. 
Such constraints could also alter natural walking behavior in both normative individuals and patients, as the presence of observers and instrumentation can induce measurable changes in gait, known as Hawthorne effects \cite{cicirelli2021human, robles2015spatiotemporal}, driving subjects to perform better rather than reproducing their everyday walking pattern. These limitations are particularly noticeable for context‑dependent or paroxysmal gait disturbances, such as freezing of gait in Parkinson’s disease or certain post‑stroke gait statuses, that are modulated by emotional, attentional, and environmental factors. In these cases, the disturbance may not manifest during brief, supervised assessments \cite{ardestani2020effect, conde2023triggers}, but representative gait behaviour could be captured through long-term monitoring, reducing contextual biases. 
Wearable IMUs partially address these constraints by enabling out-of-laboratory monitoring, yet sensor drift progressively reduces tracking accuracy and may introduce discomfort in the users over time \cite{prisco2024validity}. In this context, markerless computer vision–based motion capture has emerged as an alternative to traditional laboratory systems.
Deep learning-based human pose estimation networks can infer body joint positions from RGB video, achieving a degree of accuracy increasingly acceptable for gait analysis \cite{kanko2021concurrent, hii2023automated, wade2022applications}.
These systems drastically reduce setup time and remove the need for markers, lowering participant burden and enabling prolonged sessions and deployment outside laboratories. Within this paradigm, Real-Move\cite{realmove}, a multi-camera markerless motion-capture system, is a valid tool for capturing gait in clinical and semi-ecological environments.

Complementing these hardware developments, machine learning has become a useful tool for automated gait analysis, as deep learning models can extract and integrate features across modalities, such as video sequences, inertial sensors, and other sources \cite{khera2020role, alharthi2019deep}.
Most existing work has focused on supervised classification problems, distinguishing between normative controls and patients across different pathologies, disease stages, severity levels, and multiple gait disorder categories. Recent studies have used IMU-derived spatiotemporal features and classical machine-learning models to classify individuals with Parkinson’s disease versus normative controls and to stratify disease severity, as well as skeletal data to discriminate among primary degenerative cerebellar ataxia, hereditary spastic paraparesis, idiopathic Parkinson’s disease, and normative controls \cite{hwang2025machine, yin2024gait, martinel2024skelmamba}.
However, disease‑label classification approaches assume pathological homogeneity within diagnostic categories, which is often only partially met in clinical practice. At a global level, individuals affected by a gait disorder tend to exhibit global deviations from normative walking, such as increased spatiotemporal variability and altered gait dynamics, that easily distinguish pathological from normative locomotion \cite{moon2016gait}. A network can easily learn to classify the two groups, but this broad distinction provides limited additional information for clinical assessment or treatment planning. At a specific level, each disorder is characterized by distinct primary impairments that, in principle, provide diagnostic differentiation. In practice, however, observable gait patterns reflect both the underlying deficits and the unique compensatory strategies that each individual adopts, which are unrepresented in pre-defined diagnostic categories, often making the biomechanical manifestation of the primary deficit less noticeable.
Consequently, within the same diagnosis, gait can vary substantially across individuals, with subject‑specific compensatory strategies shaped by disease stage, affected neural systems, available motor reserves, cognitive capacity, and environmental context \cite{perry2024gait, winner2023discovering}.
Finally, common compensatory patterns observed in patients can also be seen in individuals without a neurological or orthopedic diagnosis (e.g., as transient adaptations to fatigue or minor injuries).
These factors, together with the practical challenges of collecting large, diverse, and well‑annotated pathological datasets, which is often costly both in terms of time and resources, limit the generalizability of supervised classifiers to unseen phenotypes and acquisition settings \cite{jaiswal2024benchmarking}. While augmentation can improve within‑dataset performance, synthetic variability does not necessarily reproduce the true diversity of patient‑specific gait signatures, and improvements may remain bounded by the distributions represented in training data \cite{winner2023discovering, jaiswal2024benchmarking}.

These challenges motivate a shift from disease‑label classification towards anomaly detection frameworks, where the pathological gait is modeled as a deviation from learned representations of normative locomotion \cite{pang2021deep}. Rather than assigning a diagnostic label, which collapses heterogeneous gait patterns into a single class, anomaly detection can localize deviations at the joint level, identifying which specific joints exhibit abnormal motion for each individual. This subject-specific characterization of biomechanical alterations does not rely on expectations of which joint should be impaired based on the disease, and can capture multiple subject-specific deviations without requiring them to be classified a priori as primary impairments or compensatory adaptations.
Early work has demonstrated that learning normative gait dynamics from skeleton time series can support abnormality detection without requiring exhaustive labeled pathology collections, by framing abnormal gait as a statistical deviation from normal skeletal motion and typically producing sequence-level abnormality scores \cite{nguyen2016skeleton}.
More recently, Duan et al. proposed FSGait \cite{duan2024fsgait}, a self-supervised framework enabling joint-level abnormality scoring via a normative-pose memory bank. However, memory-bank retrieval constrains the output to combinations of stored population prototypes, limiting the ability to generate individualized normative references.

To address this gap, we propose a transformer‑based model, trained exclusively on individuals without gait impairments, that generates normative gait patterns conditioned on the biomechanics and the spatio-temporal context. Transformers \cite{vaswani2017attention} are particularly well-suited for this purpose, as representing each joint-frame pair as an independent token allows self-attention to capture coordination patterns across joints that are functionally coupled but distant in the kinematic chain, learning which combinations of joints and temporal positions are most informative for generating coherent normative configurations, without locality constraints along either the temporal or the kinematic dimension. In particular, a Masked AutoEncoderTransformer \cite{he2022masked} trains a model to reconstruct missing parts of the input from the visible context. Applied to skeletal sequences, this encourages the model to learn spatial and temporal relationships among joints. Similar strategies have already been used in human-pose and skeleton-based motion analysis. P-STMO masks joints across the spatial and temporal dimensions of 2D pose sequences and reconstructs the original poses during pretraining, before using the learned representation for 3D human-pose estimation \cite{shan2022pstmo}. SkeletonMAE represents the skeleton as a graph and reconstructs masked joints and skeletal edges to learn representations for skeleton-based action recognition \cite{yan2023skeletonmae}. 

Given an observed gait sequence, potentially abnormal, anomalies are quantified as deviations from this subject‑conditioned normative baseline, enabling joint‑level localization of joints that are inconsistent with the learned normative gait patterns. By avoiding reliance on disease labels through training exclusively on normative gait and generating a continuous, individualized reference, rather than retrieving population prototypes, the proposed framework aims to improve generalization to heterogeneous and unseen gait abnormalities while providing two complementary outputs: (1) the identification of biomechanically inconsistent joints, and (2) the normative twin of the observed input, with corrected kinematic trajectories reconstructing how the observed gait would appear if the deviations were removed. 

Accordingly, the methodological objective of this study is to develop a label-free, two-pass framework that provides joint-level biomechanical inconsistency scores and an individualized normative gait reconstruction. The validation objective is to assess whether the framework preserves unseen normative gait and reduces angular deviation in simulated abnormal gait patterns.

Following validation on clinical populations, such outputs could support targeted rehabilitation planning and longitudinal monitoring of treatment response, providing clinicians with a joint-specific characterization of an individual's biomechanical gait deviations that is independent of diagnostic category and naturally accounts for individual compensatory strategies.

\section{Methods}
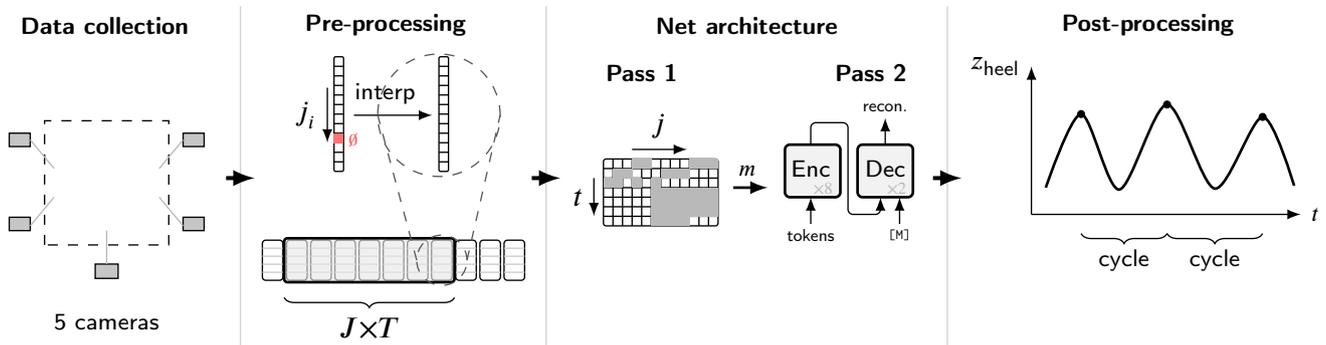
\begin{figure*}[t]
\centering
\resizebox{\linewidth}{!}{%
\begin{tikzpicture}[
  font=\footnotesize,
  >=latex,
  arrow/.style={-latex, very thick},
  lab/.style={font=\scriptsize, align=center},
  btitle/.style={font=\bfseries\scriptsize, align=center},
]

\foreach \x in {3.0, 6.2, 10.4}
  \draw[thin, gray!35] (\x,-1.65) -- (\x, 1.90);

\node[btitle] at (1.58, 1.70) {Data collection};
\node[btitle] at (4.52, 1.70) {Pre-processing};
\node[btitle] at (8.30, 1.70) {Net architecture};
\node[btitle] at (12.50,1.70) {Post-processing};

\draw[arrow] (2.85, 0.10) -- (3.15, 0.10);
\draw[arrow] (6.05, 0.10) -- (6.35, 0.10);
\draw[arrow] (10.25,0.10) -- (10.55,0.10);

\def\gs{0.1}   

\newcommand{\SingleColAt}[1]{%
  \begin{scope}[shift={#1}, x=\gs cm, y=\gs cm]
    \draw[rounded corners=0.5pt] (0,0) rectangle (1,12);
    \foreach \y in {1,...,11} \draw (0,\y) -- (1,\y);
  \end{scope}
}

\newcommand{\FillSingleColAt}[2]{%
  \begin{scope}[shift={#1}, x=\gs cm, y=\gs cm]
    \foreach \y in {#2}{
      \fill[gray!55] (0,\y) rectangle ++(1,1);
    }
  \end{scope}
}

\newcommand{\TokGridAt}[1]{%
  \begin{scope}[shift={#1}, x=\gs cm, y=\gs cm]
    \draw[rounded corners=0.8pt] (0,0) rectangle (12,7);
    \foreach \x in {1,...,11} \draw (\x,0) -- (\x,7);
    \foreach \y in {1,...,6}  \draw (0,\y) -- (12,\y);
  \end{scope}
}

\newcommand{\FillCellsAt}[2]{%
  \begin{scope}[shift={#1}, x=\gs cm, y=\gs cm]
    \foreach \y/\x in {#2}{
      \fill[gray!55] (\x,\y) rectangle ++(1,1);
    }
  \end{scope}
}

\begin{scope}[shift={(1.60, 0.05)}]
  \draw[dashed, thin] (-0.65,-0.65) rectangle (0.65,0.65);

  \fill[gray!45] (-1.02, 0.38) rectangle (-0.80, 0.52);
  \draw[thin]    (-1.02, 0.38) rectangle (-0.80, 0.52);
  \draw[thin, gray!55] (-0.80, 0.45) -- (-0.55, 0.15);

  \fill[gray!45] (-1.02,-0.50) rectangle (-0.80, -0.36);
  \draw[thin]    (-1.02,-0.50) rectangle (-0.80, -0.36);
  \draw[thin, gray!55] (-0.80,-0.45) -- (-0.55, -0.15);

  \fill[gray!45] (0.80, 0.38) rectangle (1.02, 0.52);
  \draw[thin]    (0.80, 0.38) rectangle (1.02, 0.52);
  \draw[thin, gray!55] (0.80, 0.45) -- (0.55, 0.15);

  \fill[gray!45] (0.80,-0.50) rectangle (1.02, -0.36);
  \draw[thin]    (0.80,-0.50) rectangle (1.02, -0.36);
  \draw[thin, gray!55] (0.80,-0.45) -- (0.55, -0.15);

  \fill[gray!45] (-0.12,-1.0) rectangle (0.12,-0.85);
  \draw[thin]    (-0.12,-1.0) rectangle (0.12,-0.85);
  \draw[thin, gray!55] (0,-0.85) -- (0,-0.5);

  \node[lab, font=\scriptsize] at (0, -1.45) {5 cameras};
\end{scope}

\begin{scope}[shift={(4.58, 0.05)}]

  \pgfmathsetmacro{\fw}{0.215}
  \pgfmathsetmacro{\fgap}{0.038}
  \pgfmathsetmacro{\fh}{0.42}
  \pgfmathsetmacro{\stripY}{-0.72}
  \pgfmathsetmacro{\xzero}{-1.65}

  \pgfmathsetmacro{\vecH}{12*0.048}        
  \pgfmathsetmacro{\vecW}{0.048}
  \pgfmathsetmacro{\vecY}{0.12}            
  \pgfmathsetmacro{\vecLx}{-0.72}          
  \pgfmathsetmacro{\vecRx}{-0.72+0.048+0.42} 
  \pgfmathsetmacro{\vecTop}{\vecY+12*0.048}

  \foreach \i in {0,...,10} {
    \pgfmathsetmacro{\xp}{\xzero + \i*(\fw+\fgap) + 0.3}
    \draw[thin, fill=white, rounded corners=1pt]
      (\xp, \stripY-0.3) rectangle (\xp+\fw, \stripY+\fh-0.3);
    \foreach \r in {1,2,3,4} {
      \pgfmathsetmacro{\ry}{\stripY - 0.3 + \r * \fh/5}
      \draw[thin, gray!40] (\xp, \ry) -- (\xp+\fw, \ry);
    }
  }

  \pgfmathsetmacro{\wAl}{\xzero + 1*(\fw+\fgap) - 0.025 + 0.3}
  \pgfmathsetmacro{\wAr}{\xzero + 7*(\fw+\fgap) + \fw + 0.025 + 0.3}
  \draw[thick, rounded corners=1pt, fill=gray!18, fill opacity=0.65]
    (\wAl, \stripY-0.02-0.3) rectangle (\wAr, \stripY+\fh+0.02-0.3);

  \pgfmathsetmacro{\wBl}{\xzero + 2*(\fw+\fgap) - 0.025}
  \pgfmathsetmacro{\wBr}{\xzero + 8*(\fw+\fgap) + \fw + 0.025}

  \draw[decorate, decoration={brace, mirror, amplitude=3pt}, thin]
    (\wAl, -1 + \stripY+\fh+0.10) -- (\wAr, -1+ \stripY+\fh+0.10)
    node[midway, yshift=-9pt, font=\small] {$J{\times}T$};
  
    \pgfmathsetmacro{\zfx}{\xzero + 7*(\fw+\fgap) + 0.3}
    \pgfmathsetmacro{\zfxr}{\zfx + \fw}
    
    \pgfmathsetmacro{\zfcx}{\zfx + \fw/2}
    \pgfmathsetmacro{\zfcy}{\stripY + \fh/2 - 0.3}
    \pgfmathsetmacro{\zfr}{sqrt((\fw/2+0.02)*(\fw/2+0.02) + (\fh/2+0.02)*(\fh/2+0.02))}
    
    \draw[dashed, thin, black!65] (\zfcx, \zfcy) circle (\zfr cm);

  \SingleColAt{(\vecLx+0.12, \vecY)}
  \begin{scope}[shift={(\vecLx+0.12, \vecY)}, x=\gs cm, y=\gs cm]
    \fill[red!55] (0,3) rectangle ++(1,1);
  \end{scope}
   \node[lab, font=\tiny, red!70, anchor=west]
    at (\vecLx+0.16, \vecY+0.33) {$\emptyset$};

  \draw[->] (\vecLx+0.05, \vecTop+0.3) -- (\vecLx+0.05, \vecY+0.3)
    node[midway, left, font=\small] {$j_i$};

  \pgfmathsetmacro{\midarrowY}{\vecY + 12*0.048/2}
  \pgfmathsetmacro{\arrowLx}{\vecLx + \vecW + 0.1}
  \pgfmathsetmacro{\arrowRx}{\vecRx - 0.1}
  \draw[-latex, thin]  (\arrowLx+ 0.18, \midarrowY+0.3) -- (\arrowRx+0.78, \midarrowY+0.3) node[pos=0.4, above, font=\scriptsize] {interp};

  \SingleColAt{(\vecRx+0.75, \vecY)}
  \pgfmathsetmacro{\vecRcx}{\vecRx + 0.024}

  \pgfmathsetmacro{\bothcx}{(\zfcx-0.04 + \vecRx + \vecW) / 2}
  \pgfmathsetmacro{\bothcy}{\vecY + 12*0.048/2 + 0.3}
  \pgfmathsetmacro{\halfspan}{(\vecRx + \vecW - (\zfcx-0.04)) / 2}
  \pgfmathsetmacro{\bothr}{sqrt(\halfspan*\halfspan + (12*0.048/2+0.06)*(12*0.048/2+0.06)) + 0.15}
  \draw[dashed, thin, black!65] (\vecRx+0.75, \bothcy) circle (\bothr cm);

  \draw[dashed, thin, black!65] (\zfcx-\zfr, \zfcy) -- (\vecRx+0.75-\bothr, \bothcy);
  \draw[dashed, thin, black!65] (\zfcx+\zfr, \zfcy) -- (\vecRx+0.75+\bothr, \bothcy);

\end{scope}

\begin{scope}[shift={(8.4, -0.1)}]

  \node[lab, font=\bfseries\scriptsize] at (-1.2, 1.3) {Pass 1};
  \node[lab, font=\bfseries\scriptsize] at ( 1.2, 1.3) {Pass 2};

  \TokGridAt{(-1.6, -0.3)}
  \FillCellsAt{(-1.6, -0.3)}{0/5, 0/6, 0/7, 0/8, 1/5, 1/6, 1/7, 1/8, 1/9, 1/10, 1/11, 2/5, 2/6, 2/7, 2/8, 2/9, 2/10, 2/11, 3/5, 3/6, 3/7, 3/8, 3/9, 3/10, 3/11, 4/0, 4/1, 4/3, 4/5, 5/1, 5/2, 5/6, 5/7, 5/8, 6/3, 6/4, 6/9, 6/10, 6/11}

  \draw[->] (-1.7, 0.2) -- (-1.7, -0.2)
    node[midway, left, font=\small] {$t$};
  \draw[->] (-1.3, 0.5) -- (-1.3+12*0.048, 0.5)
    node[midway, above, font=\small] {$j$};

  \draw[-latex, thick] (-0.24, 0.1) --  node[above, pos=0.4, font=\scriptsize] {$m$} (0.12, 0.1);


  \draw[thick, fill=gray!10, rounded corners=1.5pt] (0.3, 0.0) rectangle (0.88, 0.55);
  \node[lab, font=\scriptsize]          at (0.57, 0.30) {Enc};
  \node[lab, font=\tiny, gray!55] at (0.70, 0.1) {$\times 8$};

  \draw[thick, fill=gray!10, rounded corners=1.5pt]
    (1.06, 0.00) rectangle (1.64, 0.55);
  \node[lab, font=\scriptsize]          at (1.35, 0.30) {Dec};
  \node[lab, font=\tiny, gray!55] at (1.47, 0.1) {$\times 2$};

  \draw[-latex, thin] (0.57, -0.28) -- (0.57, 0.00);
  \node[lab, font=\tiny] at (0.57, -0.40) {tokens};

  \draw[-latex, thin] (1.35, 0.55) -- (1.35, 0.83);
  \node[lab, font=\tiny] at (1.35, 0.94) {recon.};

  \node[lab, font=\tiny] at (1.5, -0.40) {\texttt{[M]}};
  \draw[-latex, thin] (1.5, -0.28) -- (1.5, 0.00);

  \draw[-latex, thin, rounded corners=1.5pt]
       (0.57, 0.55)   
    -- (0.57, 0.78)  
    -- (0.96, 0.78)  
    -- (0.96, -0.18) 
    -- (1.3, -0.18) 
    -- (1.3, 0.00); 

\end{scope}
\begin{scope}[shift={(12.30, -0.1)}]

  \begin{scope}[shift={(-1.02,-0.18)}]
    \draw[->, thin] (0,0) -- (2.82,0)
      node[lab, anchor=west] {$t$};
    \draw[->, thin] (0,0) -- (0,1.55)  node[left] {$z_{\text{heel}}$};
    \draw[thick] plot[smooth] coordinates
      {(.15,.28)(.52,1.05)(.92,.26)
       (1.42,1.15)(1.92,.27)(2.42,1.02)(2.75,.30)};
    \foreach \px/\py in {.52/1.05,1.42/1.15,2.42/1.02}
      \fill (\px,\py) circle (0.042cm);
    \draw[decorate,
          decoration={brace,amplitude=3pt,mirror},thin]
      (.52,-.20) -- (1.42,-.20)
      node[midway,yshift=-8pt,lab] {cycle};
    \draw[decorate,
          decoration={brace,amplitude=3pt,mirror},thin]
      (1.42,-.20) -- (2.42,-.20)
      node[midway,yshift=-8pt,lab] {cycle};
  \end{scope}
\end{scope}

\end{tikzpicture}%
}
\caption{\textit{Pipeline overview.}
  Five cameras at 30 Hz yield 3D joint positions. Pre-processing estimates missing joints via constrained interpolation and tokenizes sequences into $J{\times}T$ joint--frame tokens using a 7-frame sliding window with stride~1. Pass~1 gives the masking pattern producing mask~$m$; Pass~2 reconstructs masked joints using a MAE Transformer. Post-processing segments gait cycles via left heel-height peak detection.}
\label{fig:pipeline}
\end{figure*}

To generate joint inconsistency scores and corrected kinematic trajectories, markerless 3D joint trajectories are pre-processed into a compact token representation and fed to the detection and correction model; gait cycles are then segmented from the output for phase-aligned evaluation (Figure \ref{fig:pipeline}).

\label{sec:methods}
\subsection{Data collection}
Data acquisition was performed using the markerless camera-based 3D motion-capture system Real-Move \cite{realmove} (Genoa, Italy), which employs deep learning pose estimation algorithms to extract three-dimensional skeletal joint positions from synchronized multi-camera video with real-time processing.

Nineteen anatomical landmarks per frame were tracked, comprising: nose, neck, left and right shoulders, left and right elbows, left and right wrists, pelvis, left and right hips, left and right knees, left and right ankles, and left and right toe tips and heels. Three-dimensional joint coordinates were extracted in a unified global reference frame following system calibration. The coordinate axes were defined as: $z$-axis (vertical, positive upward), $y$-axis (mediolateral, positive from right to left), and $x$-axis (anteroposterior, positive forward). Joint positions were expressed as Cartesian coordinates in meters relative to the world origin.

Data were collected in a controlled indoor laboratory environment with standardized lighting conditions. A linear walkway of approximately 4 meters was established, with clear markings indicating start and stop positions. Five RGB cameras, operating at 30 Hz, were used and positioned in the perimeter of the room to provide comprehensive multi-view coverage of the walking corridor.

A total of 160 adults participated in this study (age: 31.1 ± 6.2 years, range 22–57; 63 female, 97 male).
All participants were normative subjects, with no known gait pathologies, neurological or musculoskeletal disorders, or temporary injuries that would affect the walking pattern. Participants self-reported their health status, confirmed the absence of any conditions affecting locomotion during the recruitment process, and provided informed consent. The study was conducted in accordance with the ethical standards approved by the Ethics Committee of Azienda Sanitaria Locale (ASL) Genovese N.3 under Protocol IIT\_HRII\_ERGOLEAN 156/2020.
The dataset was partitioned at the participant level into a model-development cohort (\(N{=}150\) participants) and a held-out test cohort (\(N{=}10\) participants). Within the model-development cohort, 20 participants were randomly selected for validation, while the remaining 130 participants were used for training. The ten test participants were excluded from both subsets and reserved for the final evaluation.

The 150 model-development participants performed three walking trials at each of the three self-selected speed conditions: habitual walking velocity, slower, and faster. 
Speed conditions were self-selected to preserve natural gait variability without external pacing.
For each trial, participants began from a standing position at the starting mark and came to a complete stop when reaching the endpoint marker. This protocol ensured that steady-state walking was captured within the measurement zone, with acceleration and deceleration phases also being recorded. The order of speed conditions was fixed (normal, slow, fast) across all participants, ensuring that participants first established their habitual velocity as a reference before walking slower or faster.

\subsection{Pre processing}
In markerless motion-capture systems, temporary joint detection failures and self-occlusions can result in incomplete keypoint trajectories. Although the multi-camera architecture of Real-Move substantially mitigates these issues, occasional missing samples still occur. Therefore, missing joint position samples were estimated using constant‑velocity temporal interpolation, assuming each joint continued moving at the velocity observed in neighboring frames. After interpolation, the following biomechanical constraints were applied in order: (1) bone lengths were preserved by projecting the child joints onto spheres centered at the parent joint with radius equal to the child bone segment length, (2) joints constrained by two incident bone segments (like knees and elbows) were constrained to satisfy two-sphere distance requirements, and (3) a ground‑contact heuristic prevented foot markers from passing through the floor level during stance.
To remove global translation, the pelvis position was set as the coordinate system origin in each frame. The pelvis orientation was aligned with the global vertical axis to eliminate whole-body rotation, and local coordinate frames were then estimated along the fixed kinematic chain, and each joint rotation was calculated relative to the local coordinate frame of its parent joint. Joint orientations were parameterized as intrinsic $XYZ$ Euler angles in radians and, to obtain a continuous input representation, were mapped to sin–cosine pairs.
The reduced kinematic model consisted of 12 joints: neck, left and right shoulders, left and right elbows, pelvis, left and right hips, left and right knees, and left and right ankles.
Walking sequences were segmented into overlapping temporal windows of 7 frames, with stride 1. Each joint-frame pair $(j,t)$ was encoded as a token, a discrete representational unit, yielding 84 tokens per window ($J{=}12$ $\times$ $T{=}7$). Each token was represented by a 12-dimensional feature vector, $\mathbf{f}_{j,t}$, formed by concatenating the sine–cosine encoding of the intrinsic $XYZ$ Euler angles (6 values) and a 6D rotation representation, consisting of the first two orthonormal vectors of the rotation matrix derived from the same Euler angles.
This hybrid representation provided complementary views of joint rotation in both angle space (sine–cosine pairs) and geometric space (rotation matrix columns), allowing the network to learn from whichever view was more informative.

\subsection{Net architecture}

\begin{figure}
    \centering
    \tikzset{
  overshoot line to/.style={
    to path={($(\tikztostart)!-(#1)!(\tikztotarget)$)--($(\tikztotarget)!-(#1)!(\tikztostart)$)\tikztonodes}
  }
}
\scalebox{0.8}{
\begin{tikzpicture}[
    >=LaTeX, 
    very thick,
    arrow/.style={
        -latex,
        very thick,
        rounded corners=0.2cm
    },
    block/.style={
        rectangle,
        fill=gray!10,
        rounded corners=3mm,
        draw,
        very thick
    },
    layer/.style={
        rectangle,
        fill=white!10,
        rounded corners=1mm,
        inner xsep=0em,
        inner ysep=0.25em,
        minimum height=1.4em,
        align=center,
        text width=2.5cm,
        draw,
        very thick
    },
    input/.style={ 
        circle,
        minimum width=2.25em,
        draw,
        fill=gray!10,
        thick
    },
    input2/.style={ 
        rectangle,
        rounded corners=1mm,
        inner xsep=0em,
        inner ysep=0.25em,
        minimum height=1.4em,
        align=center,
        text width=2.5cm,
        draw,
        fill=gray!10,
        thick
    },
    layer2/.style={ 
        rectangle,
        rounded corners=1mm,
        inner xsep=0em,
        inner ysep=0.25em,
        minimum height=0.8cm,
        align=center,
        text width=1.5cm,
        draw,
        fill=white!10,
        thick
    },
    do path picture/.style={%
        path picture={%
          \pgfpointdiff{\pgfpointanchor{path picture bounding box}{south west}}%
            {\pgfpointanchor{path picture bounding box}{north east}}%
          \pgfgetlastxy\x\y%
          \tikzset{x=\x/2,y=\y/2}%
          #1
        }
    },
    sin wave/.style={do path picture={    
        \draw [line cap=round] (-3/4,0)
        sin (-3/8,1/2) cos (0,0) sin (3/8,-1/2) cos (3/4,0);
        }
    }
]

\tikzset{
  emb/.style={
    path picture={
      \pgfpointdiff{\pgfpointanchor{path picture bounding box}{south west}}
                   {\pgfpointanchor{path picture bounding box}{north east}}
      \pgfgetlastxy\x\y
      \begin{scope}[shift={(path picture bounding box.center)}, x=\x/2, y=\y/2]
        \draw (-0.55,-0.45) -- (-0.55,0.45);
        \draw (-0.25,-0.20) -- (-0.25,0.45);
        \draw ( 0.05,-0.45) -- ( 0.05,0.45);
        \draw ( 0.35,-0.10) -- ( 0.35,0.45);
        \draw ( 0.55,-0.45) -- ( 0.55,0.45);
      \end{scope}
    }
  }
}

\newcommand{\SkeletonIconAt}[2]{%
  \begin{scope}[shift={#1}, x=0.05cm, y=0.05cm]
    \draw[line width=0.45pt] (5,13)--(5,10);
    \draw[line width=0.45pt, draw=gray!75] (3,13)--(7,13);
    \draw[line width=0.45pt, draw=gray!75] (3,13)--(1,9);
    \draw[line width=0.45pt, draw=gray!75] (7,13)--(9,9);
    \draw[line width=0.45pt] (4,10)--(3.5,5);
    \draw[line width=0.45pt] (6,10)--(6.5,5);
    \draw[line width=0.45pt] (4,10)--(6,10);
    \draw[line width=0.45pt] (3.5,5)--(4,1);
    \draw[line width=0.45pt] (6.5,5)--(6,1);
    \draw[line width=0.45pt, draw=gray!75] (4,1)--(3,0.5);
    \draw[line width=0.45pt, draw=gray!75] (4,1)--(4.3,0.5);
    \draw[line width=0.45pt, draw=gray!75] (6,1)--(7,0.5);
    \draw[line width=0.45pt, draw=gray!75] (6,1)--(5.7,0.5);
    \draw[line width=0.45pt, draw=gray!75] (5,16)--(5,13);
    \foreach \jx/\jy in {5/10,4/10,6/10,3.5/5,6.5/5,5/13}
      {\fill (\jx,\jy) circle (0.38);}
    \foreach \jx/\jy in {5/16,2/11,8/11,4.3/0.5,5.7/0.5,3/0.5,7/0.5,4/1,6/1,3/13,7/13,1/9,9/9}
      {\fill[gray!95] (\jx,\jy) circle (0.38);}
    \def\highlightjoint{#2}%
    \def\nojoint{}%
    \ifx\highlightjoint\nojoint\else
      \fill[red] #2 circle (0.38);
    \fi
  \end{scope}
}

\newcommand{\SkeletonFanAt}[1]{%
  \pgfmathsetmacro{\FanDX}{0.5}
  \coordinate (SF6) at ($(#1)+(6*\FanDX cm, 0)$);
  \coordinate (SF5) at ($(#1)+(5*\FanDX cm, 0)$);
  \coordinate (SF4) at ($(#1)+(4*\FanDX cm, 0)$);
  \coordinate (SF3) at ($(#1)+(3*\FanDX cm, 0)$);
  \coordinate (SF2) at ($(#1)+(2*\FanDX cm, 0)$);
  \coordinate (SF1) at ($(#1)+(1*\FanDX cm, 0)$);
  \coordinate (SF0) at ($(#1)+(0*\FanDX cm, 0)$);
  \SkeletonIconAt{(SF6)}{}
  \SkeletonIconAt{(SF5)}{}
  \SkeletonIconAt{(SF4)}{}
  \SkeletonIconAt{(SF3)}{}
  \SkeletonIconAt{(SF2)}{}
  \SkeletonIconAt{(SF1)}{}
  \SkeletonIconAt{(SF0)}{}
}

    \node[] (iemb) at (1.25,-1.8){};

\pgfmathsetmacro{\SkW}{0.60}   
\pgfmathsetmacro{\SkH}{0.9}    
\pgfmathsetmacro{\Odx}{0.13}   
\pgfmathsetmacro{\Ody}{0.04}   

\begin{scope}[shift={([xshift=-0.65cm, yshift=-0.75cm]iemb.south)}]
  \foreach \k in {0,1,2,3,4,5,6}{
    \draw[thick, fill=white]
      ({\k*\Odx}, {-\k*\Ody}) rectangle ({\k*\Odx+\SkW}, {-\k*\Ody+\SkH});
  }
  \SkeletonIconAt{({(6*\Odx)+0.05}, {(-6*\Ody)++0.025})}{}
\end{scope}

    \node[] (oemb) at (5.25, 0) {};

    \node[layer, above=1em of iemb] (linear_enc) {Linear};

    \node[layer, above=2em of oemb] (ass) {Token\\Assembly};

    \node[circle, draw, minimum size=0.25em, inner sep=0pt, above=2.5em of linear_enc] (sum1) {$\mathbf{+}$};
    \node[circle, draw, minimum size=0.25em, inner sep=0pt, above=2.5em of ass] (sum2) {$\mathbf{+}$};
    \node [circle, draw, sin wave, minimum size=2em, left=0.8em of sum1] (pe1) {};
    \node [circle, draw, sin wave, minimum size=2em, right=0.8em of sum2] (pe2) {};

    \node [circle, draw, emb, minimum size=2em, right=0.8em of sum1] (ee1) {};
    \node [circle, draw, emb, minimum size=2em, left=0.8em of sum2] (ee2) {};

     \node[layer] at (ass.west -| sum1.center)  (tm) {Token \\ Masking};

    \node[layer2, dashed, left=3.3em of tm] (pass1) {PASS 1};
    \draw[arrow, dashed] (pass1) -- node[midway, above] {$m$} (tm);

    \node[layer] (add1) at (1.25,5.35) {Add \& Norm};
    \node[layer] (attn1) at (1.25,4.5) {Multi-Head \vspace{-0.05cm} \linebreak Attention};
    \draw[] (attn1) -- (add1);
    \node[layer] (add4) at (1.25,7.9) {Add \& Norm};
    \node[layer] (ff1) at (1.25,7.05) {Feed \vspace{-0.05cm} \linebreak Forward};
    \draw[] (ff1) -- (add4);

    \node[layer] (add2) at (5.25,5.35) {Add \& Norm};
    \node[layer] (attn2) at (5.25,4.5) {Multi-Head \vspace{-0.05cm} \linebreak Attention};
    \draw[] (attn2) -- (add2);
    \node[layer] (add5) at (5.25,7.9) {Add \& Norm};
    \node[layer] (ff2) at (5.25,7.05) {Feed \vspace{-0.05cm} \linebreak Forward};
    \draw[] (ff2) -- (add5);

    \node[circle, draw,minimum size=0.25em, inner sep=0pt, above=2.5em of add4] (m) {$\setminus$};
    \draw[] (add4) -- node[pos=0.67, left, align=center]  {memory} (m);

    \draw[arrow,dashed] (pass1.west) -- ($(pass1.west) + (-0.4,0)$) |- ($(m.west) + (0,0)$) node[pos=0.75, above, align=center] {$m$} ;    

    \coordinate (d1) at ($(attn2.south east) + (0.75,-0.7)$);
    \coordinate (d2) at ($(add5.north west) + (-0.15,0.05)$);
    \coordinate (e1) at ($(attn1.south east) + (0.15,-0.7)$);
    \coordinate (e2) at ($(add4.north west) + (-0.75,0.05)$);
    \begin{scope}[on background layer]
         \node[block, fit=(d1) (d2)] (decoder) {};
         \node[block, fit=(e1) (e2)] (encoder) {};
    \end{scope}    

    \node[layer, above=1.5em of add5] (linear) {Linear};
    \node[above =1em of linear] (probs) {};
    \node[above left=1em of linear] (prob) {};

    \begin{scope}[shift={([xshift=0.95cm, yshift=0.3 cm]prob)}]
  \foreach \k in {0,1,2,3,4,5,6}{
    \draw[thick, fill=white]
      ({\k*\Odx}, {-\k*\Ody}) rectangle ({\k*\Odx+\SkW}, {-\k*\Ody+\SkH});
  }
  \SkeletonIconAt{({(6*\Odx)+0.05}, {(-6*\Ody)++0.025})}{}
\end{scope}

    \draw[arrow] (add1) -- (ff1);
    \draw[arrow] (add2) -- (ff2);
    \draw[arrow] (add5) -- (linear);
    \draw[arrow] (iemb) -- (linear_enc);
    \draw[arrow] (linear_enc) -- (sum1);
    \draw[arrow] (sum1) -- (tm);
    \draw[arrow] (tm) -- node[pos=0.33, left, align=center]  {visible tokens\\$+$\\\texttt{[MASK]}} (attn1);
    \draw[arrow] (sum2) -- (attn2);
    \draw[] (sum1) -- (pe1);
    \draw[] (sum2) -- (pe2);
    \draw[] (ee1) -- (sum1);
    \draw[] (ee2) -- (sum2);

    \draw[arrow] (ff1.south)++(0, -0.6) -| ($(add4.west) + (-0.6,-0.5)$) |- (add4.west);
    \draw[arrow] (attn1.south)++(0, -0.6) -| ($(add1.west) + (-0.6,-0.5)$) |- (add1.west);

    \draw[arrow] (attn2.south)++(0, -0.6) -| ($(add2.east) + (0.6,-0.5)$) |- (add2.east);
    \draw[arrow] (ff2.south)++(0, -0.6) -| ($(add5.east) + (0.6,-0.5)$) |- (add5.east);

    \draw[arrow] (attn1.south)++(0, -0.4) -| ($(attn1.south) + (-1,0)$);
    \draw[arrow] (attn1.south)++(0, -0.4) -| ($(attn1.south) + (1,0)$);
   
    \draw[arrow] (m.north) |- ($(m.north) + (1,0.3)$) -| ($(m.north) + (2,-1)$) |- ($(oemb.south) + (-0.5,0.3)$) node[pos=0.75, below, align=center] {visible\\latents} -- ($(ass.south)+ (-0.5,0)$);

    \node[] at ($(ass.south)+ (0.5,-1)$) (mask) {\texttt{[MASK]}};
    \draw[arrow] (mask) -- ($(ass.south)+ (0.5,0)$);
    \draw[arrow] (ass) -- (sum2);
    
    \node[] at ($(pe1.west) + (-0.3,0)$) {$\mathbf{P}$};
    \node[] at ($(pe2.east) + (0.3,0)$) {$\mathbf{P}$};
    \node[] at ($(ee1.east) + (0.3,0)$) {$\mathbf{E}$};
    \node[] at ($(ee2.west) + (-0.3,0)$) {$\mathbf{E}$};

    \draw[arrow] (linear) -- (probs);

    \node[anchor=east] at ($(encoder.south east) + (0,0.2)$) {$8\times$};
    \node[anchor=west] at ($(decoder.south west) + (0.1,0.2)$) {$2\times$};
    
\coordinate (decFront) at ($ (ee1.east) + (3mm,-2mm) $);
\coordinate (decBack)  at ($ (ee2.east) + (3mm,-2mm) + (4mm,4mm) $);

\end{tikzpicture}
}
    \caption{\textit{Masked autoencoder Transformer for joint reconstruction (Pass 2).}
    A 7-frame window is tokenized into joint–frame tokens and linearly projected, then indexed by a sinusoidal positional code $\mathbf{P}$ and three learned embeddings $\mathbf{E}$ (joint type, frame index, and motion/velocity). 
    A mask pattern $m$ (provided by Pass~1) specifies which token positions are hidden. The \textbf{Token Masking} operator replaces the selected tokens with a learned \texttt{[MASK]} placeholder, yielding a fixed-length masked sequence (visible tokens + \texttt{[MASK]}) processed by an 8-layer Transformer encoder.
    The encoder outputs a full-length \emph{memory} sequence; visible memory vectors are selected using $\emph{memory}$$\setminus$$m$ and injected into the decoder input via \textbf{Token Assembly}, while masked positions are filled with \texttt{[MASK]}; $\mathbf{P}$ and $\mathbf{E}$ are added again before the 2-layer decoder. The Transformer decoder reconstructs the full token window, from which the reconstructed last-frame tokens are retained as the corrected pose estimate at inference.}
    \label{fig:pass2}
\end{figure}
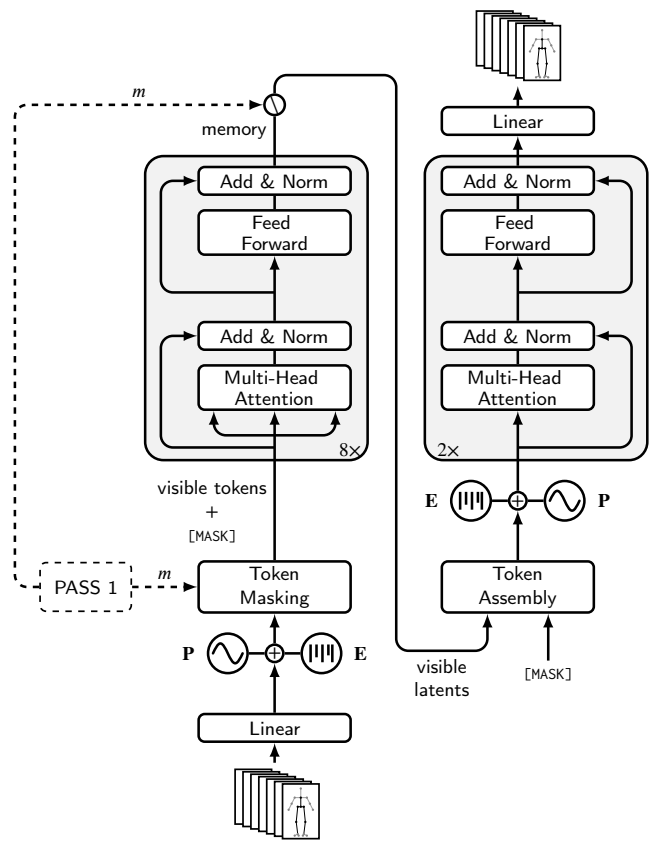
\begin{figure*}
\centering
\resizebox{\linewidth}{!}{%
\begin{tikzpicture}[
  font=\footnotesize,
  >=latex,
  arrow/.style={->, thick},
  sep/.style={dashed, thick},
  outer/.style={rounded corners=2pt, fill=white, draw, line width=0.7pt},
  box/.style={rounded corners=2pt, fill=white, draw, line width=0.7pt},
  outbox/.style={rounded corners=2pt, fill=white, draw, line width=0.7pt},
  title/.style={font=\bfseries}
]

\def\Wtot{16.0}
\def\Htot{9.6}
\def\Gap{0.55}
\def\Wpanel{7.7}
\def\Hbox{2.00}
\def\Vgap{0.38}

\pgfmathsetmacro{\HalfW}{0.5*\Wtot}
\pgfmathsetmacro{\SepX}{\Wpanel+\Gap}
\pgfmathsetmacro{\BB}{\Htot}
\pgfmathsetmacro{\MaskY}{-(\Htot+0.45)}
\pgfmathsetmacro{\PassTwoY}{-(\Htot+0.95)}

\path[use as bounding box] (0,1.4) rectangle (\Wtot,{-\BB});

\draw[sep]   (\SepX,0) -- (\SepX,-\Htot+0.8);

\node[anchor=north west, title] at (0.0,1) {\normalsize(A) Training mode};
\node[anchor=north west, title] at (\Wpanel+\Gap+0.03,1) {\normalsize (B) Inference mode};

\def\gs{0.30} 
\def\im{0.2}

\newcommand{\TokGridAt}[1]{%
  \begin{scope}[shift={#1}, x=\gs cm, y=\gs cm]
    \draw[rounded corners=0.8pt] (0,0) rectangle (12,7);
    \foreach \x in {1,...,11} \draw (\x,0) -- (\x,7);
    \foreach \y in {1,...,6}  \draw (0,\y) -- (12,\y);
  \end{scope}
}

\newcommand{\FillCellsAt}[2]{
  \begin{scope}[shift={#1}, x=\gs cm, y=\gs cm]
    \foreach \y/\x in {#2}{
      \fill[gray!55] (\x,\y) rectangle ++(1,1);
    }
  \end{scope}
}

\newcommand{\SingleRowAt}[1]{%
  \begin{scope}[shift={#1}, x=\im cm, y=\im cm]
    \draw[rounded corners=0.8pt] (0,0) rectangle (12,1);
    \foreach \x in {1,...,11} \draw (\x,0) -- (\x,1);
  \end{scope}
}

\newcommand{\FillSingleRowAt}[2]{%
  \begin{scope}[shift={#1}, x=\im cm, y=\im cm]
    \foreach \x/\dummy in {#2}{
      \fill[gray!55] (\x,0) rectangle ++(1,1);
    }
  \end{scope}
}

\newcommand{\SingleColAt}[1]{%
  \begin{scope}[shift={#1}, x=\gs cm, y=\gs cm]
    \draw[rounded corners=0.8pt] (0,0) rectangle (1,12);
    \foreach \y in {1,...,11} \draw (0,\y) -- (1,\y);
  \end{scope}
}

\newcommand{\FillSingleColAt}[2]{%
  \begin{scope}[shift={#1}, x=\gs cm, y=\gs cm]
    \foreach \y in {#2}{
      \fill[gray!55] (0,\y) rectangle ++(1,1);
    }
  \end{scope}
}

\newcommand{\CurriculumPlotAt}[1]{%
  \begin{scope}[shift={#1}, x=0.020cm, y=0.733cm]
    \fill[gray!55]
      (0,0)--(25,0.02)--(120,0.90)--(160,0.90)--(160,0)--cycle;
    \fill[gray!15]
      (0,1)--(0,0)--(25,0.02)--(120,0.90)--(160,0.90)--(160,1)--cycle;
    
    \draw[thin] (0,0) rectangle (160,1);
    
    \foreach \xp/\lbl in {0/0, 25/5, 120/60}{
      \draw[thin] (\xp,-0.06)--(\xp,0);
      \node[font=\small, anchor=north] at (\xp,-0.07) {\lbl};
    }
    \draw[thin] (160,-0.06)--(160,0);
    \node[font=\small, anchor=north] at (160,-0.07) {250};
    \node[font=\small, anchor=north] at (140,-0.05) {$\cdots$};
    
    \node[font=\normalsize, anchor=north] at (80,-0.52) {epoch};

    \node[font=\normalsize, anchor=west, gray] at (7, 0.7) {random};
    \node[font=\normalsize, anchor=west, black!75] at (80, 0.4) {structured};

    \draw[thin] (-1,0.10)--(0,0.10);
    \draw[thin] (-1,0.90)--(0,0.90);
    \node[font=\small, anchor=east] at (-1.5,0.10) {10\%};
    \node[font=\small, anchor=east] at (-1.5,0.90) {90\%};
  \end{scope}
}
\def\Jarrow{0.28}
\coordinate (RandSW)   at (0.8, -3);
\coordinate (StructSW) at (7.2, -3);

\newcommand{\SkeletonIconAt}[2]{%
  \begin{scope}[shift={#1}, x=0.1cm, y=0.1cm] 
    \draw[line width=0.45pt] (5,13)--(5,10);
    \draw[line width=0.45pt, draw=gray!75] (3,13)--(7,13);
    \draw[line width=0.45pt, draw=gray!75] (3,13)--(1,9);  \draw[line width=0.45pt, draw=gray!75] (7,13)--(9,9);
    \draw[line width=0.45pt] (4,10)--(3.5,5);  \draw[line width=0.45pt] (6,10)--(6.5,5);
    \draw[line width=0.45pt] (4,10)--(6,10);
    \draw[line width=0.45pt] (3.5,5)--(4,1);   \draw[line width=0.45pt] (6.5,5)--(6,1);
    \draw[line width=0.45pt,draw=gray!75] (4,1)--(3,0.5);   \draw[line width=0.45pt, draw=gray!75] (4,1)--(4.3,0.5);
    \draw[line width=0.45pt,draw=gray!75] (6,1)--(7,0.5);   \draw[line width=0.45pt,draw=gray!75] (6,1)--(5.7,0.5);
    \draw[line width=0.45pt,draw=gray!75] (5,16)--(5,13);
    
    \foreach \jx/\jy in {5/10,4/10,6/10,3.5/5,6.5/5,5/13}
      {\fill (\jx,\jy) circle (0.38);}
      \foreach \jx/\jy in {5/16,2/11,8/11,4.3/0.5,5.7/0.5,3/0.5,7/0.5,4/1,6/1,3/13,7/13,1/9,9/9}
      {\fill[gray!95]  (\jx,\jy) circle (0.38);}

    \def\highlightjoint{#2}%
    \def\nojoint{}%
    \ifx\highlightjoint\nojoint\else
      \fill[red] #2 circle (0.38);
    \fi
  \end{scope}
}

\newcommand{\SkeletonFanAt}[1]{%
  \pgfmathsetmacro{\FanDY}{0.25} 
  \pgfmathsetmacro{\Ang}{30}     
  \pgfmathsetmacro{\FanDX}{\FanDY/tan(\Ang)} 

  \coordinate (SF5) at ($(#1)+(5*\FanDX cm, 5*\FanDY cm)$);
  \coordinate (SF4) at ($(#1)+(4*\FanDX cm, 4*\FanDY cm)$);
  \coordinate (SF3) at ($(#1)+(3*\FanDX cm, 3*\FanDY cm)$);
  \coordinate (SF2) at ($(#1)+(2*\FanDX cm, 2*\FanDY cm)$);
  \coordinate (SF1) at ($(#1)+(1*\FanDX cm, 1*\FanDY cm)$);
  \coordinate (SF0) at ($(#1)+(0*\FanDX cm, 0*\FanDY cm)$);

  \SkeletonIconAt{(SF5)}{(3.5,5)}
  \SkeletonIconAt{(SF4)}{(6.5,5)}
  \SkeletonIconAt{(SF3)}{(4,10)}
  \SkeletonIconAt{(SF2)}{(6,10)}
  \SkeletonIconAt{(SF1)}{(5,10)}
  \SkeletonIconAt{(SF0)}{(5,13)}
}

\def\gsi{0.08} 
\pgfmathsetmacro{\TileW}{12*\gsi}
\pgfmathsetmacro{\TileH}{7*\gsi}
\pgfmathsetmacro{\HalfTileW}{0.5*\TileW}
\pgfmathsetmacro{\TileTopY}{\TileH+0.06}
\pgfmathsetmacro{\TileRowYOffset}{\TileH+0.20}
\def\TileGap{0.08} 

\newcommand{\TokGridSmallAt}[1]{%
  \begin{scope}[shift={#1}, x=\gsi cm, y=\gsi cm]
    \draw[rounded corners=0.6pt, black!80] (0,0) rectangle (12,7);
    \foreach \x in {1,...,11} \draw[black!50] (\x,0)--(\x,7);
    \foreach \y in {1,...,6}  \draw[black!50] (0,\y)--(12,\y);
  \end{scope}
}

\newcommand{\MaskColSmallAt}[2]{
  \begin{scope}[shift={#1}, x=\gsi cm, y=\gsi cm]
    \fill[gray!95] (#2,0) rectangle ++(1,7);
  \end{scope}
}

\newcommand{\BjPlotAt}[1]{%
  \begin{scope}[shift={#1}, x=0.268cm, y=1.22cm]

    \draw[->] (0,0) -- (13.2,0) node[below, font=\normalsize] {$t$};
    \draw[->] (0,0) -- (0,1.8)  node[left, font=\normalsize] at (-0.5,1.7) {$B_j(t)$};

    \draw[dashed, thin, black!55] (0,0.5) -- (12,0.5);
    \node[font=\normalsize, anchor=west] at (12.1,0.5) {$\tau$};

    \draw[gray!70, line width=0.55pt]
      plot[smooth] coordinates
        {(0,.09)(2,.10)(4,.27)(6,.11)(8,.10)(10,.10)(12,.09)};
    \fill[gray!70] (4,.27) circle (0.18);

    \draw[gray!70, line width=0.55pt]
      plot[smooth] coordinates
        {(0,.10)(3,.11)(5,.24)(7,.12)(9,.10)(12,.10)};
    \fill[gray!70] (5,.24) circle (0.18);

    \draw[gray!70, line width=0.55pt]
      plot[smooth] coordinates
        {(0,.11)(2,.12)(6,.31)(8,.13)(10,.11)(12,.11)};
    \fill[gray!70] (6,.31) circle (0.18);

    \draw[gray!70, line width=0.55pt]
      plot[smooth] coordinates
        {(0,.08)(4,.09)(7,.26)(9,.10)(11,.09)(12,.08)};
    \fill[gray!70] (7,.26) circle (0.18);

    \draw[red!72, line width=0.70pt]
      plot[smooth] coordinates
        {(0,.09)(2,.10)(3,.50)(4,.76)(5,.53)(7,.12)(9,.10)(12,.09)};
    \fill[red!72] (4,.76) circle (0.23);

    \draw[red!72, line width=0.70pt]
      plot[smooth] coordinates
        {(0,.10)(1,.11)(3,.45)(5,1.14)(6,.49)(8,.12)(10,.10)(12,.10)};
    \fill[red!72] (5,1.14) circle (0.23);
    \node[font=\normalsize, red!80, anchor=south west, inner sep=0.5pt]
      at (5.15,1.3) {$\mathbf{\tilde{B}_j}$};

    \draw[red!72,  line width=0.65pt] (9.1,1.7)--(10.3,1.7);
    \node[font=\normalsize, anchor=west] at (10.4,1.7) {flagged};
    \draw[gray!70, line width=0.65pt] (9.1,1.44)--(10.3,1.44);
    \node[font=\normalsize, anchor=west] at (10.4,1.44) {stable};

  \end{scope}
}


\pgfmathsetmacro{\Rxr}{\Gap-0.3}

\node[anchor=north west, minimum width=1.2cm,
      minimum height=2.8cm] (Lrand) at (0.2,-1.1) {}; 
\SingleColAt{($(Lrand.center)+(-0.11,-1.32)$)}
\FillSingleColAt{($(Lrand.center)+(-0.11,-1.32)$)}{1,3,5,8,10}
\node[font=\normalsize] at ($(Lrand.south)+(0.1, -0.3)$) {random};
\node[font=\small]   at ($(Lrand.north)+(0,1.2)$) {$t_i$};
\draw[->] ($(RandSW)+(-\Jarrow, 2.34)$)
       -- ($(RandSW)+(-\Jarrow, -0.3)$)
  node[midway, left, font=\normalsize] {$j_i$};

\node[anchor=north west, minimum width=1.2cm,
      minimum height=2.8cm] (Lstruct) at (6.5,-1.1) {};
\SingleColAt{($(Lstruct.center)+(-0.11,-1.32)$)}
\FillSingleColAt{($(Lstruct.center)+(-0.11,-1.32)$)}{7,8,9}
\node[font=\normalsize] at ($(Lstruct.south)+(0.1, -0.3)$) {structured};
\node[font=\small]   at ($(Lstruct.north)+(0,1.2)$) {$t_i$};
\draw[->] ($(StructSW)+(\Jarrow, 2.34)$)
       -- ($(StructSW)+(\Jarrow, -0.3)$)
  node[midway, right, font=\normalsize] {$j_i$};
  
\draw[->, line width=1.5pt]
  ($(RandSW)+(0.5, 1.22)$) -- ($(StructSW)+(-0.5, 1.22)$);
  
\node[anchor=north, minimum width=\Wpanel cm, minimum height=1.0cm] (Lcurr)
  at ($(Lrand.center)!0.5!(Lstruct.center)+(0.2,0.3)$) {};
\CurriculumPlotAt{($(Lcurr.center)+(-1.52,-0.26)$)}

\node[anchor=north west, minimum width=\Wpanel cm, minimum height=\Hbox cm] (Lspan)
  at (-0.95, -6.5) {};
\node[anchor=south, font=\normalsize]
  at ($(Lspan.south)+(1.17,-0.10)$) (Ltemp) {temporal coherence};
\TokGridAt{($(Lspan.center)+(-0.72,-0.50)$)}
\FillCellsAt{($(Lspan.center)+(-0.72,-0.50)$)}{0/5, 0/6, 0/7, 0/8, 1/5, 1/6, 1/7, 1/8, 1/9, 1/10, 1/11, 2/5, 2/6, 2/7, 2/8, 2/9, 2/10, 2/11, 3/5, 3/6, 3/7, 3/8, 3/9, 3/10, 3/11, 4/0, 4/1, 4/3, 4/5, 5/1, 5/2, 5/6, 5/7, 5/8, 6/3, 6/4, 6/9, 6/10, 6/11}
\draw[thick]  ($(Lspan.center)+(-0.85,-0.08)$) -- ($(Lspan.center)+(-0.85,0.52)$);
\node[anchor=east, font=\small] at ($(Lspan.center)+(-0.92,0.22)$) {span $\ell$};
\draw[->]($(Ltemp.center)+(-1.1,2.55)$)
       -- ($(Ltemp.center)+(1.3, 2.55)$)
  node[midway, above, font=\normalsize] {$j_i$};

\draw[->] ($(Lspan.center)+(3,0.9)$)
       -- ($(Lspan.center)+(3, -0.3)$)
  node[midway, right, font=\normalsize] {$t_i$};

\draw[decorate, decoration={brace, mirror, raise=14pt, amplitude=9pt}] ($(Lrand.south west) + (0.1,-0.4)$) -- ($(Lstruct.south east) + (0.25,-0.4)$);

\pgfmathsetmacro{\Rx}{\Wpanel+\Gap+0.25}
\pgfmathsetmacro{\Rcx}{\Rx+0.5*\Wpanel}

\node[anchor=north west, minimum width=\Wpanel cm, minimum height=1.6cm] (Rtiles)
  at (\Rx, 0) {};

\coordinate (Tile0SW) at ($(Rtiles.north west)+(0,-\TileRowYOffset)$);

\TokGridSmallAt{(Tile0SW)}
\node[font=\normalsize, anchor=south] at ($(Tile0SW)+(\HalfTileW,\TileTopY)$) {base};

\pgfmathsetmacro{\TileLabelY}{\TileH+0.16}

\foreach \k/\col/\name in {1/0/neck, 2/5/pelvis, 3/6/L.hip, 4/7/L.knee, 5/9/R.hip, 6/10/R.knee}{
  \pgfmathsetmacro{\tx}{\k*(\TileW+\TileGap)}
  \coordinate (Tile\k SW) at ($(Tile0SW)+(\tx,0)$);
  \TokGridSmallAt{(Tile\k SW)}
  \MaskColSmallAt{(Tile\k SW)}{\col}
  \node[font=\normalsize, anchor=base, inner sep=0pt]
    at ($(Tile\k SW)+(\HalfTileW,\TileLabelY)$) {\name};
}

\coordinate (Tile0B) at ($(Tile0SW)+(\HalfTileW,0)$);
\foreach \k in {1,...,6}{
  \coordinate (Tile\k B) at ($(Tile\k SW)+(\HalfTileW,0)$);
}

\node[outbox, anchor=north, minimum width=7 cm, minimum height=0.72cm] (Rpass2band)
  at (\Rcx-0.22, -1.4) {\large Pass 2};

\draw[->, thin] (Tile0B) -- ($(Tile0B |- Rpass2band.north)$) node[midway, right, font=\small] {$m$};
\foreach \k in {1,...,6}{
  \draw[->, thin] (Tile\k B) -- ($(Tile\k B |- Rpass2band.north)$)
  node[midway, right, font=\small] {$m$};
}

\draw[->, thin]  ($(Rpass2band.south)$) -- ($(Rpass2band.south)+(0,-0.5)$);
\coordinate (RbaseSW) at ($(Rpass2band.south)+(-2,-2.075)$);
\SkeletonIconAt{(RbaseSW)}{}
\node[font=\normalsize, anchor=north] at ($(RbaseSW)+(0.55,-0.15)$) {base};

\coordinate (RfanSW) at ($(Rpass2band.south)+(0.3,-3)$);
\SkeletonFanAt{RfanSW}
\node[font=\normalsize, anchor=north] at ($(RfanSW)+(2.3, 0.5)$) {tiles};

\node[font=\large, anchor=north] at ($(Rpass2band.south)+(0,-0.85)$) {VS};
\draw[->, thin] ($(Rpass2band.south)+(0,-3.2)$)
             -- ($(Rpass2band.south)+(0,-3.80)$)
  node[font=\normalsize] at ($(Rpass2band.south)+(0.25,-3.45)$)
    {$\forall t$};

\node[anchor=north, minimum width=3.9cm, minimum height=2.7cm, inner sep=3pt] (Rplot)
  at ($(Rpass2band.south)+(0,-3.80)$) {};

\BjPlotAt{($(Rplot.south west)+(0.22,0.26)$)}

\node[outbox] (P2)    at (\HalfW + \Gap/2, -\Htot) {\large Pass 2};

\draw[arrow] (Rplot.south) |- (P2.east) node[pos=0.8, below, font=\normalsize]
    {$m$};

\draw[arrow] ($(Ltemp.south)+ (-0.08,0)$) |- (P2.west) node[pos=0.8, below, font=\normalsize]
    {$m$};

\end{tikzpicture}%
}
\caption{\textit{Pass 1: mask identification.} Training uses a curriculum of synthetic masks (random \(\rightarrow\) structured) with temporally coherent spans to produce the mask list \(m\). Inference uses tiled occlusions and a badness score \(B_j\) to select unreliable joints and produce \(m\). In both cases \(m\) is then inputted to Pass~2 for masked reconstruction.}
\label{fig:pass1}
\end{figure*}
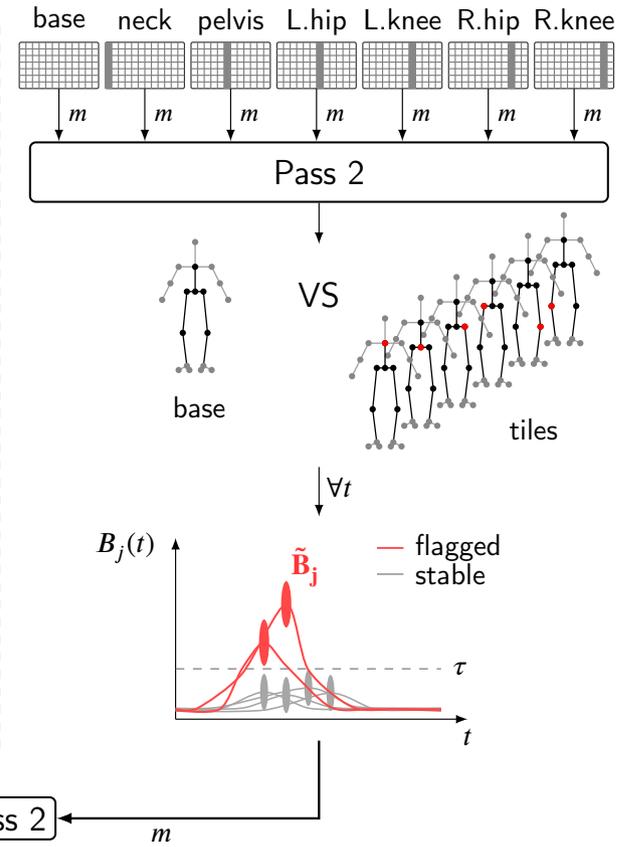 

Pose correction was modeled as a masked spatiotemporal reconstruction problem, with a focus on the last frame of the observation window. The framework, based on a masked autoencoder (MAE) Transformer trained exclusively on normative gait data, leverages redundancy across joints and time to generate plausible joint configurations from partial observations. Indeed, selected joints were masked and their configurations inferred from the remaining spatiotemporal context, which spans the concurrent poses of visible joints at the last frame and their full temporal evolution across the window, forcing the model to learn joint recovery from partial observations, which is then directly exploited at inference for targeted kinematic correction. The core net architecture, shown in Figure \ref{fig:pass2}, comprises a Transformer encoder and a lightweight Transformer decoder, operating on a fixed-length token window. This architecture supports generative tasks, with the encoder summarizing spatiotemporal context into a latent representation and the decoder reconstructing the complete token window.

At the network input, the joint-frame token sequence is linearly projected to the model embedding dimension. The projected tokens are then indexed with a fixed sinusoidal positional encoding, \textbf{P}, and augmented with three learned embeddings, \textbf{E}: a joint-type embedding (identifying which joint), a frame embedding (temporal position within the window), and a motion embedding (instantaneous angular velocity).
The encoder, which consists of 8 Transformer layers, processes a $J \times T$ masked version of the token sequence with masked token positions replaced by learned \texttt{[MASK]} placeholder tokens. Its output is a $J \times T$ latent \emph{memory} sequence, from which only visible memory vectors are selected, by excluding the masked indices, $m$, from the memory sequence, $\emph{memory}$$\setminus$$m$. 
Then, at the decoder input, the token sequence is constructed by merging the $J \times T$ learned \texttt{[MASK]} placeholder token sequence with the visible encoder memory vectors at their corresponding indices. The same positional encoding, \textbf{P}, and learned embeddings, \textbf{E}, are then added. The decoder, which consists of 2 Transformer layers, reconstructs the full token window, from which the reconstructed last-frame tokens are retained as the corrected pose estimate at inference.
Both encoder and decoder layers consist of multi-head self-attention followed by a feed-forward network, with residual connections and layer normalization applied throughout. The number of attention heads was set to 12, matching the number of joints in the kinematic model; given this choice, a head dimension of 24 was selected as appropriate for effective attention mechanisms, yielding the 288-dimensional latent embedding space ($12 \times 24{=}288$). The feed-forward network had an internal dimension of 1536 units in both the encoder and decoder layers. Dropout of 0.1 was applied throughout the network. For the 84-token windows \((L=84)\), the self-attention term within each Transformer layer has computational complexity \(O(L^2d)\) because it evaluates pairwise interactions among the tokens, whereas the projection and feed-forward terms scale linearly with \(L\) for fixed model dimensions. Since the token length is fixed, the trial-level computational cost scales linearly with the number of processed windows.

During model development, alternative temporal-window lengths (one, three, and seven frames), rotation representations, numbers of attention heads, and encoder--decoder depths were explored through a sequential model-selection procedure in which each candidate modification was evaluated separately. Decisions on whether to retain each modification were based exclusively on reconstruction losses on the normative validation data. This ensured that, like parameter learning, architecture selection relied solely on normative gait, while the simulated abnormal-gait data remained reserved for final testing. After the architecture and training hyperparameters had been fixed, the training and validation subsets were merged, and the final model was trained using data from all 150 model-development participants.

During training, the model was optimized in a self-supervised inpainting framework, aiming to reconstruct the $J \times T$ token window while making the network robust to missing inputs. Training proceeded using the AdamW optimizer with learning rate $2 \times 10^{-4}$, momentum parameters $\beta_1{=}0.9$ and $\beta_2{=}0.95$, and weight decay $5 \times 10^{-2}$ were applied to all parameters except bias terms and normalization layers. The learning rate remained constant throughout training, and no learning-rate scheduler was used. Gradients were clipped to a maximum norm of 1.0. The model was trained for 250 epochs with batch size 256.

Mask patterns are generated by Pass~1 in Figure~\ref{fig:pass1}, which operates in two distinct modes depending on whether the model is being trained or used at inference.

In training mode, Figure \ref{fig:pass1}(A), Pass 1 implemented a curriculum of synthetic masks designed to improve robustness to missing inputs. The masking strategy evolved progressively over training epochs, following a curriculum that shifted from random joint dropout (50\% masked) in the early epochs to structured spatial patterns mimicking biomechanically plausible occlusions (e.g., full limb removal) in the later epochs, with the transition completed by epoch 60. For each training batch, the probability of selecting a structured spatial mask was \(5\%\) during epochs 0--4 and \(10\%\) during epochs 5--19. It then increased linearly from \(10\%\) at epoch 20 to \(90\%\) at epoch 60 and remained \(90\%\) thereafter. The complementary probability was used for random masking.
Temporal coherence in the masking was also introduced by reusing the same joint mask across contiguous temporal spans: for each window, a span length was randomly selected with a given probability (25\% probability of a single frame span, 25\% full window, 50\% between 2 and 6 consecutive frames), and the start frame of that contiguous span was also sampled uniformly at random within the window. The same joint keep-set persisted within that span, while frames outside were sampled independently. 
This net training regime was chosen to force the model to learn redundancy across joints and frames, inferring a joint from the rest of the body and its dynamics, and the model was optimized using a combination of five complementary reconstruction losses, which were summed directly, with a coefficient of 1 for each term: (i) sine-cosine L1 reconstruction at the final frame, (ii) masked and (iii) visible joint reconstruction over the full sequence, (iv) angular velocity consistency, and (v) a context-invariance loss penalizing inconsistent reconstructions under different mask patterns.
All losses operated on sine-cosine angle representations to ensure angular continuity across the $[-\pi, \pi]$ boundary.

At inference time, since the corruption pattern is unknown, randomly masking joints would be counterproductive, as it could discard information from reliable joints while retaining corrupted ones.  Therefore, Pass 1 switches from synthetic masking, improving robustness, to an anomaly-driven mask estimator that identifies biomechanically inconsistent joints. Pass 2, then apply the inferred mask to selectively remove only the flagged joints for targeted correction.
In inference mode, Figure \ref{fig:pass1}(B), Pass 1 operates in tiled mode, where the input window is replicated 6 times. In each tile $j$, the encoder is prevented from attending to the tokens of joint $j$, while keeping the rest of the spatiotemporal context visible, obtaining a set of alternative reconstructions. The 6 joints (neck, pelvis, hips, knees) were selected to capture the essential degrees of freedom for gait analysis. Ankles were excluded due to their lower tracking reliability, as more prone to positional noise. An additional unmasked tile produces a baseline reconstruction, which serves as a network-normalized reference. For each joint and frame, the pipeline compares the baseline and tile predictions via forward kinematics, computing a time-varying badness score \(B_j(t)\) that quantifies geometric deviation between the resulting bone vectors, weighted by range-of-motion and degree-of-freedom sensitivity. Comparing tile and baseline outputs rather than output and raw input, as in standard anomaly detection approaches, removes two sources of bias: (i) global reconstruction artefacts shared by tile and baseline cancel out, and (ii) input artefacts are not compared directly with the model output, although they may still influence both reconstructions. The score therefore provides a joint-specific counterfactual measure of how the predicted kinematics change when evidence from that joint is withheld. The rationale is that joints whose predicted motion under targeted masking diverges significantly from the baseline, which closely tracks the input, indicate a large deviation between the observed motion and the learned normative prior. Thus, they are flagged as biomechanically inconsistent with the learned prior and the remaining spatiotemporal context.

The score comprises two components: a geometric term measuring spatial inconsistency between baseline and tile bone vectors, and a biomechanical term weighting angular deviations by their functional significance.

The biomechanical component \( C_{\text{ROM}} \) normalizes the angular difference between baseline and tile reconstructions using fixed joint-specific range-of-motion limits, weighted by the functional relevance of each rotation axis during gait:
\begin{align}
    &|\Delta \varphi_i| = |\text{wrap}_\pi (\varphi_i^\text{tile} -
        \varphi_i^\text{base})| &\in& [0,\pi] \\
    &C_{\text{ROM}} = \sum_{i \in \{x,y,z\}} w_i \cdot
        \left[\frac{|\Delta \varphi_i|}{\text{ROM}_i}\right]_0^1
        &\in& [0,1].
\end{align}
Here, \(\Delta\varphi_i\) is the wrapped angular difference for rotation axis \(i\), \(\text{ROM}_i\) is the range-of-motion normalization limit for that axis, and \(w_i\) is its degree-of-freedom importance weight, with \(\sum_i w_i=1\). The quantities \(\varphi_i^\text{tile}\) and \(\varphi_i^\text{base}\) denote the joint angles obtained from the masked-tile and unmasked-baseline reconstructions, respectively. The wrapping operator is defined as \(\text{wrap}_\pi(a)=((a+\pi)\bmod 2\pi)-\pi\), and \([a]_0^1=\min(1,\max(0,a))\) denotes clipping to \([0,1]\).

The geometric component \(E_{\text{geom}}\) measures the directional mismatch between the bone vectors produced by the baseline and tile reconstructions:
\begin{align}
    &\cos(\vartheta) =
        \frac{\mathbf{v}_\text{base}\cdot\mathbf{v}_\text{tile}}
             {\|\mathbf{v}_\text{base}\|\,\|\mathbf{v}_\text{tile}\|}
        &\in&[-1,1] \\
    &E_{\text{geom}} = \frac{1-\cos(\vartheta)}{2}
        &\in&[0,1].
\end{align}
Here, \(\vartheta\) is the angle between the baseline and tile bone vectors. \(E_{\text{geom}}\approx0\) indicates aligned bone vectors and \(E_{\text{geom}}\approx1\) indicates antiparallel directions, while \(\mathbf{v}_\text{base}\) and \(\mathbf{v}_\text{tile}\) are the parent-to-child bone vectors obtained from the baseline and tile reconstructions, respectively.

\(E_\text{geom}\) and \(C_{\text{ROM}}\) are combined into a per-joint, per-frame badness score \(B_j(t)\), which is then summarized over the trial via a peak statistic:
\begin{align} 
&B_j(t) = E_{\text{geom}} \cdot (0.5 + 0.5 \cdot C_{\text{ROM}}) &\in& [0,1] \\ 
&\tilde{B}_j =Q_{0.99}\left(\{B_j(t)\}_{t=1}^{N_w}\right) &\in& [0,1] 
\end{align}
where \(N_w\) is the number of windows in the trial and \(Q_{0.99}\) denotes the 99th percentile. The index \(j\) identifies the joint being tested, while \(t\) indexes the overlapping window and its corresponding reconstructed output frame.

The weighting factor \( (0.5 + 0.5 \cdot C_{\text{ROM}}) \) ensures that geometric error is always penalized (minimum $0.5\times$ weight) while large biomechanically meaningful deviations amplify the penalty up to $1.0\times$, preventing small angular changes from masking spatially inconsistent reconstructions. Accordingly, \(E_{\text{geom}}\) defines the underlying geometric inconsistency, whereas \(C_{\text{ROM}}\) only modulates its weight according to the relative angular deviation. $B_j(t)$ is evaluated over the trial and summarized into $\tilde B_j$ via a peak statistic, so transient corruptions accumulate into a large score while biomechanically consistent joints remain near zero. 

After processing all windows of the trial, every eligible joint satisfying \(\tilde B_j>\tau\) is selected for correction, where \(\tau\) denotes the selection threshold. Because the joint scores must be aggregated across the trial before the correction mask is defined, the current GenGait implementation operates offline at the trial level.

Pass~2 re-runs the same network, Figure \ref{fig:pass2}, with only those flagged joints masked for the encoder, forcing the model to reconstruct the current-frame pose from the remaining reliable joints and temporal dynamics, ensuring correction is applied only where evidence indicates a deviation.

\subsection{Post processing}
Post-processing was applied uniformly to training, test, and reconstructed trials, normalizing gait trials to a common temporal reference for cycle-averaged analysis, and enabling direct comparison between original and reconstructed joint trajectories on a phase-aligned basis. Hence, continuous walking trials were segmented into individual gait cycles using an automated detection algorithm that smoothed the left heel vertical position with a Savitzky-Golay filter (window=7 frames, polynomial order=2), and an energy-based activity detection identified the steady-state walking region by computing the root mean square (RMS) envelope of vertical heel velocity with a 0.5-second moving window. Gait cycle boundaries were then identified by detecting local maxima in the smoothed left heel vertical trajectory within the active region, using a prominence threshold of 0.002 m and a minimum inter-peak distance of \(0.35f_s\) samples, where \(f_s=30\)~Hz, to ensure physiological plausibility. Successive left-heel maxima, approximately corresponding to left-foot mid-swing and right-foot stance, defined the cycle start and end frames, yielding one complete cycle per interval.

\section{Experiments}
\label{sec:experiments}

The experimental evaluation addressed two complementary validation objectives: normative preservation and deviation correction.
Normative preservation evaluates the model's ability to leave normative gait intact, as a model trained on normative gait should leave biomechanically valid patterns unchanged, even when applied to previously unseen normative trials. This proves that the inpainting mechanism does not introduce deviations in the normative input kinematics.
Deviation correction evaluates the model's ability to detect and correct deviations in pathological trials, as a model trained on normative gait should identify and reduce deviations in the pathological input kinematics.

\subsection{Experimental protocol}
To assess normative preservation, the model was applied to the normative walking trials of the 10 held-out test participants, whose gait patterns were not represented during training. 
To assess deviation correction, the same 10 test participants were instructed to mimic seven distinct gait abnormalities commonly observed in neurological and orthopedic disorders. Each participant performed two iterations per anomaly, yielding 140 simulated abnormal-gait trials (10 subjects × 7 anomalies × 2 iterations). The instructed tasks, displayed in Figure \ref{fig:skel}, were:

\begin{enumerate}
    \item Circumduction (CD): exaggerated lateral swing of the leg during swing phase, compensating for insufficient hip or knee flexion on the paretic side, characteristic of hemiparetic gait;
    \item Hip hike (HH): unilateral elevation of the pelvis and hip on the swing side to clear the foot of the affected limb, typically observed in patients with ankle dorsiflexion weakness;
    \item High-steppage (HS): excessive hip and knee flexion of the affected limb during swing to compensate for foot drop, common in peroneal nerve palsy;
    \item Geriatric gait (GG): shortened stride, reduced speed, and widened base of support;
    \item Trunk extension (TE): posterior trunk lean during stance, a bilateral sagittal postural deviation, often compensating for hip extensor weakness or forward instability;
    \item Trunk flexion (TF): anterior trunk lean, a bilateral sagittal postural deviation characteristic of Parkinson's disease or a unilateral compensatory strategy for knee extensor weakness in the loading phase;
    \item Lateral trunk lean (TL): excessive mediolateral trunk displacement during stance, typically a unilateral deviation towards the affected stance limb compensating for hip abductor weakness (Trendelenburg gait), though a contralateral lean can also occur, as well as a bilateral alternation.
\end{enumerate}

Participants received verbal instructions and visual demonstrations for each pattern but did not receive biomechanical feedback. As normative individuals without prior experience mimicking gait anomalies, the resulting variability in execution reflected both inter-individual differences in motor interpretation and the natural heterogeneity observed in real compensatory strategies.

Correction was applied to every eligible joint whose aggregated Pass~1 score exceeded the threshold \((\tau=0.01)\). This value was empirically selected by computing the per-joint 99th percentiles of the Pass~1 scores obtained from the normative training trials, whose scores were generally of the order of \(10^{-3}\). Because the relationship between \(B_j(t)\) and the bone-vector angle is nonlinear, a score of \(0.01\) corresponds to an approximately \(11.5\degree\)--\(16.3\degree\) directional difference between the baseline and tile bone vectors, depending on \(C_{\text{ROM}}\), rather than to 1\% of the angular range.

\subsection{Data analysis}
Normative preservation and deviation correction were evaluated on the subset of joints that underwent the anomaly detection and reconstruction pipeline (neck, pelvis, hips, and knees, Section~\ref{sec:methods}). Within this set, only the angle directions that mostly reflect the degrees of freedom functionally relevant in gait were considered: pelvis flexion/extension, hip abduction/adduction, hip flexion/extension, and knee flexion/extension. These, indeed, correspond to the principal axes of motion during walking and, critically, to the planes in which the gait deviations primarily manifest. Rotational axes with negligible gait-phase variation (e.g. axial rotation) were excluded to avoid inflating the comparison with directions that carry no discriminative signal in the present anomaly set.
Of these, only the right-side angles are reported, as participants were instructed to perform all anomalies on the right leg, and posturally symmetric deviations manifest primarily at the pelvis, making the left and right limb responses equivalent by construction.
 
To establish a quantitative reference for normative gait, a normative band was constructed from the training dataset normal-speed walking trials (N=150 participants, three trials each). Each gait cycle was normalized to 100 frames by linear interpolation over the normalized time $[0,1]$, ensuring phase alignment across cycles of varying duration. For each joint angle \(\varphi_i(t)\), the temporal waveform was first unwrapped to remove phase discontinuities, interpolated, and then re-wrapped to the interval \([ -\pi, \pi]\) to maintain angular continuity. The normative band for each angle was defined as:
\begin{equation}
    \qquad \qquad \qquad \mu_i(t) \pm k \cdot \sigma_i(t)
\end{equation}
where \(\mu_i(t)\) is the mean angle across all training cycles at normalized time $t$, \(\sigma_i(t)\) is the standard deviation, and \(k=2.0\) defines the bandwidth.

Model performance was quantified using root-mean-square error (RMSE) to assess angular deviation from the normative kinematics. Evaluation was performed at the participant-condition level. For each combination of test participant and instructed task, gait cycles were first normalized to 100 frames, then averaged to obtain a single representative mean trajectory per joint angle, \(\bar{\varphi}_i^{\text{test}}(f)\). This mean trajectory was then compared against the training normative reference. The angular deviation magnitude, expressed in degrees, was computed frame by frame, on the wrapped angles, and then averaged over the total number of (\(F=100\)) normalized frames:
\begin{equation}
    \text{RMSE}_i = \sqrt{\frac{1}{F}\sum_{f=1}^{F} \left(\bar{\varphi}_i^{\text{test}}(f) - \mu_i(f)\right)^2}.
\end{equation}

Here, \(i\) indexes the analyzed joint angle, \(f\) indexes the normalized gait-cycle frame, \(\bar{\varphi}_i^{\text{test}}(f)\) is the participant--task mean trajectory being evaluated, and \(\mu_i(f)\) is the corresponding normative mean trajectory.

RMSE captures the magnitude of deviation from normative kinematics, with lower values indicating closer adherence to the normative mean. For normative gait validation, low RMSE in both original and reconstructed cycles would confirm that the model preserves normative kinematics without introducing artificial deviations. In simulated abnormal-gait trials, the RMSE of the original cycles was expected to be elevated for joint angles deviating from the normative reference, whereas it was expected to be lower in the reconstructed ones.

Paired statistical tests, performed separately for simulated abnormal-gait and normative trials, compared original versus reconstructed RMSE values to assess reconstruction quality.

For normative gait validation, equivalence testing assessed whether reconstructed normative kinematics remained statistically equivalent to the original across 10 participant pairs (10 participants × 1 normative condition). Normality of the paired RMSE differences was assessed using the Shapiro-Wilk test (\(\alpha {=} 0.05\)), and although the Shapiro-Wilk test did not reject normality, the small sample size ($N{=}10$) motivated the use of a non-parametric bootstrap equivalence test. Bootstrap resampling (20,000 iterations) was used to construct 90\% confidence intervals (CI) for the mean paired RMSE difference. Equivalence margins were conservatively defined as \(\delta{=}1.5 \degree\) for all joint angles, within the expected measurement noise of the markerless vision-based capture system, and representing negligible deviations. Equivalence was concluded if the 90\% CI of the mean difference fell entirely within \([-\delta, +\delta]\).

For pathological gait, paired difference testing compared original and reconstructed RMSE values across the 70 participant-anomaly combinations (10 participants × 7 anomalies). As the normality assumption was rejected by the Shapiro-Wilk test (\(\alpha {=} 0.05\)), the non-parametric Wilcoxon signed-rank test was applied to evaluate whether reconstruction significantly reduced the pathological data deviation from normative kinematics.
Effect size was quantified using rank-biserial correlation ($r_{rb}$), which ranges from -1 to +1. Negative values indicate RMSE reduction (improvement), as differences were computed as reconstructed minus original. Values of $|r_{rb}| \geq 0.5$ indicate large effects, with $r_{rb} \leq -0.5$ representing large improvement and $r_{rb} \geq +0.5$ representing large worsening.

All statistical tests were corrected for multiple comparisons using the Holm-Bonferroni procedure to control family-wise error rate at \(\alpha{=}0.05\) across the four primary joint angles analyzed (pelvis flexion/extension, right hip abduction/adduction, right hip flexion/extension, right knee flexion/extension).

To assess whether the aggregate correction effects depended on any individual participant, a leave-one-participant-out sensitivity analysis was performed. The analysis was repeated ten times, each time excluding all seven participant--task pairs associated with one participant and retaining the remaining 63 pairs. For each iteration, the Wilcoxon signed-rank tests comparing original and reconstructed RMSE were repeated, Holm-Bonferroni correction was applied across the four analyzed joint angles, and the rank-biserial correlation and percentage of participant--task pairs showing an RMSE reduction were recorded. Because every participant completed the same seven instructed tasks, each iteration retained the same task composition among the remaining participants.

\section{Results}
\label{sec:results}

\begin{figure*}[!t]
    \centering
    \includegraphics[width=\linewidth]{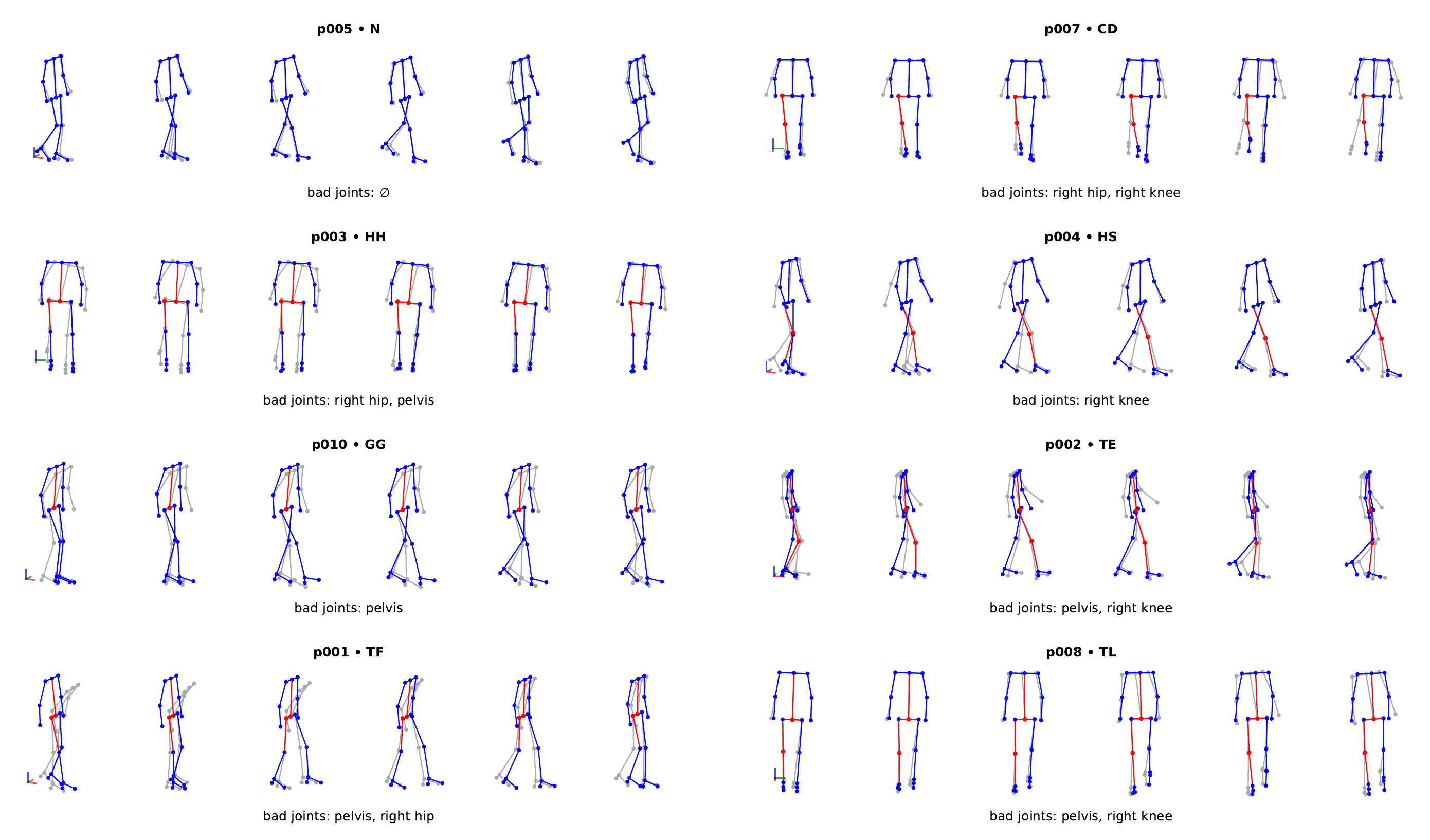}
    \caption{Skeletal reconstruction for the normative trial and the seven simulated anomalies. Each panel shows a different participant. Joints flagged as biomechanically inconsistent by Pass 1 are highlighted in red. Reconstructed skeletons (blue) are overlaid with input ones (gray). Participant IDs and flagged joints are labeled above each panel. Video animations for all conditions are available at \href{https://youtu.be/Rcm3jqR5pN4}{\texttt{https://youtu.be/GenGait}}.
    }
    \label{fig:skel}
\end{figure*}

Figure \ref{fig:skel} presents representative skeletal reconstructions for each of the seven simulated anomalies across selected participants, plus the normative trial. Joints flagged as biomechanically inconsistent by the Pass 1 anomaly detector are highlighted in red. Reconstructed skeletons are in blue, while the input skeletons, acquired by Real-Move, are in gray. 

In the representative normative trial (N), shown in Figure \ref{fig:skel}, no joint is flagged, and the reconstructed skeleton overlaps with the input one. Across all ten normative trials, 59 of the 60 joint--trial scores remained below the \(0.01\) Pass~1 threshold. The only threshold exceedance involved the left knee of participant~7, yielding an operational false-selection rate of \(1/60\), or \(1.7\%\).
For simulated abnormal-gait tasks, the selected joints are biomechanically compatible with the dominant deviations visible in the representative executions: the right hip and knee are selected for circumduction (CD), the pelvis and right hip for hip hike (HH), and the right knee for high-steppage (HS), while the pelvis is selected for the trunk-deviation tasks (TF/TE/TL) and geriatric gait (GG). In some examples, additional selected joints are compatible with mechanically coupled movements along the kinematic chain. 

\begin{figure*}[!t]
    \centering
    \includegraphics[width=\linewidth]{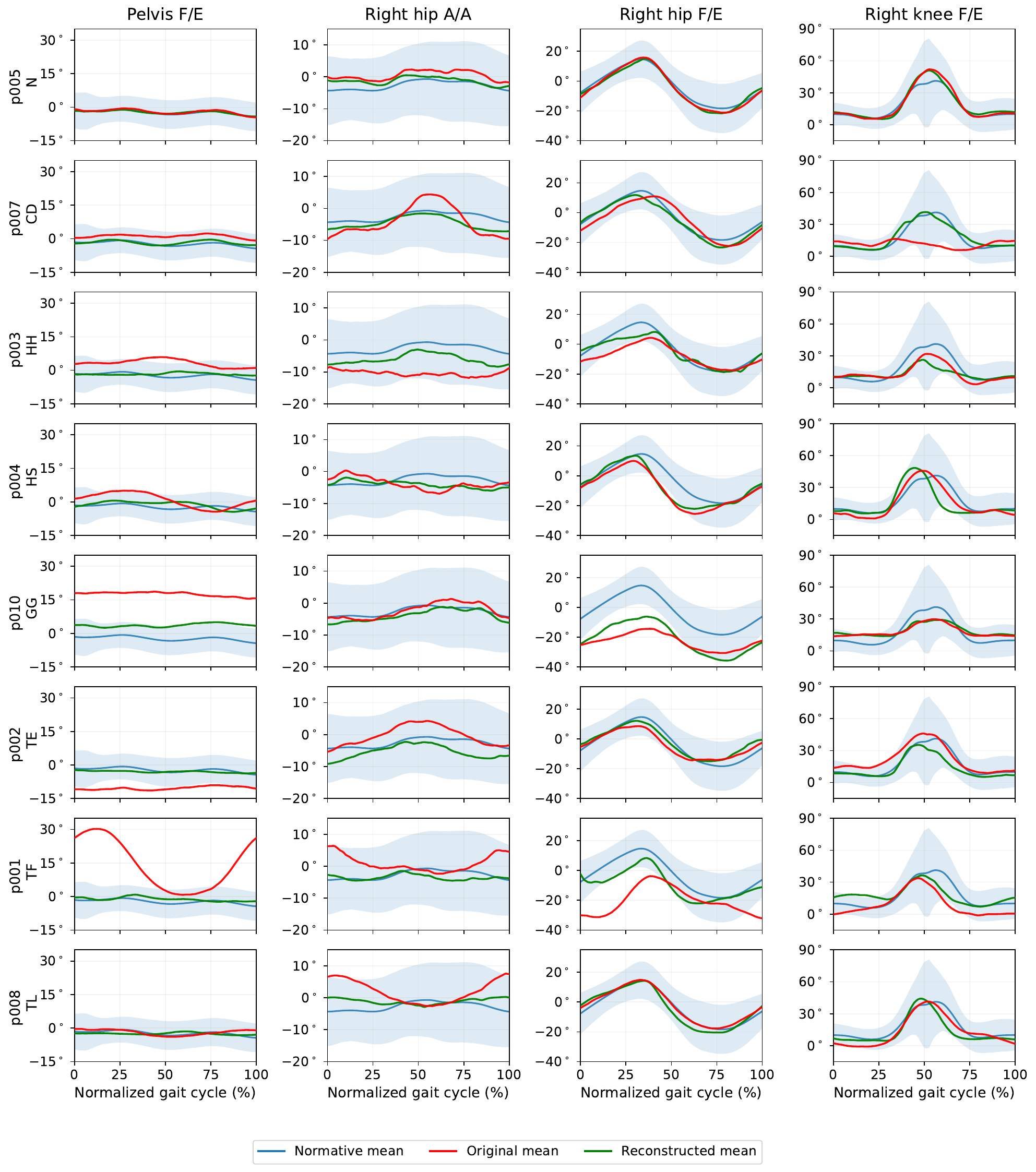}
    \caption{Joint-angle trajectories across the normalized gait cycle for the normative trial and representative executions of the seven simulated abnormal-gait tasks. The panels show pelvis flexion/extension (F/E), right hip abduction/adduction (A/A), right hip F/E, and right knee F/E for selected participant--task pairs. Blue trajectories and shaded bands represent the normative reference (\(\mu\pm2\sigma\)), original mean trajectories are shown in red, and reconstructed mean trajectories are shown in green. The horizontal axis represents the normalized gait cycle. The boundaries at 0\% and 100\% approximately delimit one right heel-strike-to-right-heel-strike cycle.}
    \label{fig:grid}
\end{figure*}

Figure \ref{fig:grid} displays normalized joint angle trajectories (pelvis flexion/extension, right hip abduction/adduction, right hip flexion/extension, right knee flexion/extension) for both the normative trial and the seven simulated abnormal-gait tasks (in red), alongside their reconstructions (in green) and the normative mean and reference band (mean $\pm 2\sigma$ from training data, in blue). In the normative trial, N, original and reconstructed trajectories, respectively, remain closely overlapping and consistently within the reference band. In the representative simulated abnormal-gait trials, original trajectories deviate from the normative envelope, particularly during phases where the instructed movement is visible (e.g., exaggerated pelvis rotation in CD, excessive hip extension during swing in TF), while reconstructed waveforms consistently shift toward the normative band in most cases while preserving the temporal phase structure of the gait cycle.

Table \ref{tab:rmse} presents RMSE values measuring angular deviation from the normative reference (training mean) for both original trials and their reconstructions, across the four primary joint angles. The first row reports the mean and inter-participant standard deviation over all normative walking trials ($N{=}10$ participants); the following rows each correspond to a representative participant selected for that task (same subjects shown in Figure \ref{fig:skel} and Figure \ref{fig:grid}). 

For normative trials, similar RMSE values between original and reconstruction indicate gait preservation. Original and reconstructed RMSE values remain nearly identical and consistently low ($<5\degree$ across all joints), with mean differences of $<0.30\degree$ (pelvis flexion/extension), $<0.32\degree$ (hip abduction/adduction), $<0.82\degree$  (hip flexion/extension), and $<0.25\degree$ (knee flexion/extension).

For the simulated abnormal-gait trials, lower RMSE indicates closer agreement with normative kinematics. Hence, a reconstruction RMSE smaller than the original one confirms that the framework moved the reconstructed gait toward the normative range. 
Reconstruction consistently reduced RMSE across most task--joint combinations. The largest corrections were observed in examples involving sagittal-plane deviations: trunk flexion (TF) showed a reduction from $19.49\degree$ to $1.39\degree$ in pelvis flexion and $19.81\degree$ to $7.61\degree$ in hip flexion; geriatric gait (GG) improved from $19.85\degree$ to $6.02\degree$ in pelvis flexion/extension. Circumduction (CD) exhibited substantial knee flexion correction ($14.39\degree$ to $3.32\degree$), consistent with the reduced knee flexion during the swing phase.
Some task--joint combinations in the representative examples showed minimal change or increases in RMSE. For example, high-steppage (HS) knee flexion increased from $5.39\degree$ to $10.65\degree$, and hip-hike (HH) knee flexion increased from $7.03\degree$ to $9.35\degree$.

\begin{table}[t]
\centering
\begingroup
\caption{RMSE (degrees) between observed gait and the normative reference for original simulated abnormal-gait trials and their reconstructions, across four joint angles. The normative row reports mean \(\pm\) standard deviation across the 10 held-out normative participants. Each subsequent row corresponds to one representative participant selected for that instructed task.}
\label{tab:rmse}

\setlength{\tabcolsep}{3pt}

\resizebox{\columnwidth}{!}{%
\begin{tabular}{@{}ll
    >{\columncolor{gray!15}}c c
    >{\columncolor{gray!15}}c c
    >{\columncolor{gray!15}}c c
    >{\columncolor{gray!15}}c c@{}}
\toprule
\textbf{Participant} & \textbf{Trial} & 
\multicolumn{2}{c}{\textbf{Pelvis flex/ext}} & 
\multicolumn{2}{c}{\textbf{Hip abd/add}} & 
\multicolumn{2}{c}{\textbf{Hip flex/ext}} & 
\multicolumn{2}{c}{\textbf{Knee flex/ext}} \\
\cmidrule(lr){3-4} 
\cmidrule(lr){5-6} 
\cmidrule(lr){7-8} 
\cmidrule(lr){9-10}
 & & Orig. & Recon. & Orig. & Recon. & 
     Orig. & Recon. & Orig. & Recon. \\
\midrule
mean \(\pm\) SD & N
& \(1.80\pm1.48\)
& \(1.50\pm1.03\)
& \(2.71\pm1.26\)
& \(2.39\pm0.88\)
& \(4.62\pm2.01\)
& \(3.80\pm1.69\)
& \(3.91\pm1.04\)
& \(4.16\pm0.83\) \\
p007 & CD & 3.47  & 0.87 & 3.81  & 2.35 & 4.65  & 2.92  & 14.39 & 3.32  \\
p003 & HH & 5.81  & 1.28 & 8.07  & 3.50 & 7.01  & 3.81  & 7.03  & 9.35  \\
p004 & HS & 4.21  & 1.71 & 3.28  & 2.44 & 8.07  & 7.10  & 5.39  & 10.65 \\
p010 & GG & 19.85 & 6.02 & 1.29  & 1.50 & 20.51 & 17.53 & 7.35  & 7.37  \\
p002 & TE & 8.23  & 1.09 & 3.18  & 3.15 & 4.70  & 4.04 & 6.32  & 6.05  \\
p001 & TF & 19.49 & 1.39 & 5.05  & 1.66 & 19.81 & 7.61 & 11.09 & 8.00  \\
p008 & TL & 1.21  & 1.06 & 6.53  & 2.51 & 1.69  & 3.43 & 5.53  & 6.86  \\
\bottomrule
\end{tabular}%
}

\endgroup
\end{table}

\begin{figure*}[t]
    \centering
    \subfloat[Normative gait: RMSE equivalence testing between original and reconstructed trials (N=10 participants). Bootstrap 90\% confidence intervals demonstrate statistical equivalence within $\delta{=}1.5\degree$ margins for all joints.]{%
        \includegraphics[width=\linewidth]{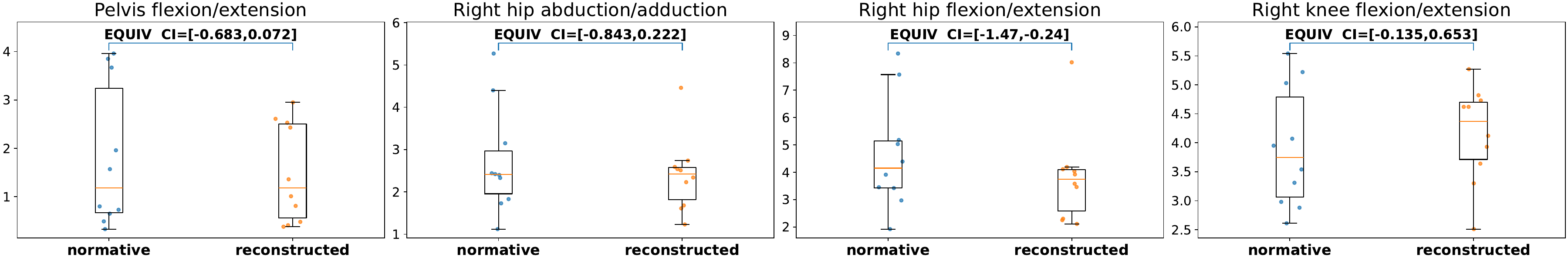}%
        \label{fig:n}%
    }\\[1em]
    \subfloat[Pathological gait: RMSE comparison between original and reconstructed trials across 70 participant-anomaly pairs (10 participants × 7 anomalies). Wilcoxon signed-rank tests with Holm-Bonferroni correction: *** indicates $p < 0.001$ for all four joint angles.]{%
        \includegraphics[width=\linewidth]{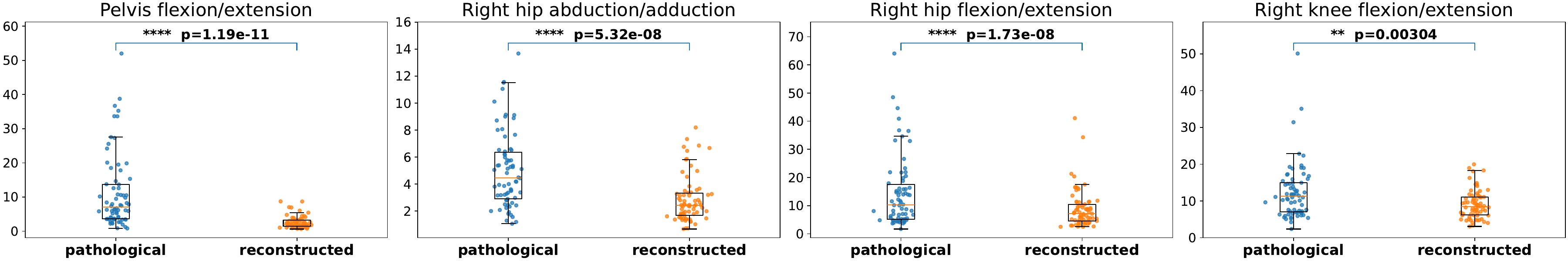}%
        \label{fig:non-n}%
    }
    \caption{Statistical validation of reconstruction performance. (a)Normative trials. (b) Pathological trials.}
    \label{fig:boxplots}
\end{figure*}

Figure \ref{fig:n} shows RMSE distributions for original versus reconstructed normative walking trials (N=10 participants). Bootstrap equivalence testing with $\delta{=}1.5\degree$ margins confirmed statistical equivalence for all four joint angles. The 90\% CI for mean RMSE difference fell entirely within $[-1.5\degree, +1.5\degree]$: pelvis flexion/extension $\text{CI}{=}[-0.68\degree, 0.07\degree]$, hip abduction/adduction $\text{CI}{=}[-0.84\degree, 0.22\degree]$, hip flexion/extension $\text{CI}{=}[-1.47\degree, -0.24\degree]$, and knee flexion/extension $\text{CI}{=}[-0.14\degree,\allowbreak\, 0.65\degree]$. All comparisons achieved statistical equivalence after Holm-Bonferroni correction (pelvis flexion/extension: $p_{equiv} < 0.001$; hip abduction/adduction: $p_{equiv} < 0.001$; hip flexion/extension: $p_{equiv}{=}0.043$; knee flexion/extension: $p_{equiv} < 0.001$).
Mean RMSE differences between original and reconstructed normative trials were small: $-0.30\degree$ (pelvis flexion/extension), $-0.32\degree$ (hip abduction/adduction), $-0.82\degree$ (hip flexion/extension), and $+0.24\degree$ (knee flexion/extension). All differences fell well below the threshold of $1.5\degree$.

Figure \ref{fig:non-n} presents RMSE distributions for original versus reconstructed simulated abnormal-gait cycles across all 70 participant--task pairs. Results are aggregated across all tasks and participants to assess the model's general correction capability across heterogeneous gait deviations. This aggregate analysis evaluates overall correction across the heterogeneous dataset rather than correction performance for each instructed task separately.

Reconstruction significantly reduced angular deviation from the normative reference for all four analyzed joint angles (pelvis flexion/extension, right hip abduction/adduction, right hip flexion/extension, right knee flexion/extension). Wilcoxon signed-rank tests with Holm-Bonferroni correction yielded $p < 0.001$ for pelvis flexion/extension ($p{=}1.19 \times 10^{-11}$), hip abduction/adduction ($p{=}5.32 \times 10^{-8}$), hip flexion/extension ($p{=}1.73 \times 10^{-8}$), and $p{=}0.003$ for knee flexion/extension. Rank-biserial effect sizes were large for all joints: $r_{rb}{=}-0.96$ (pelvis flexion/extension), $r_{rb}{=}-0.76$ (hip abduction/adduction), $r_{rb}{=}-0.80$ (hip flexion/extension), and $r_{rb}{=}-0.41$ (knee flexion/extension), with negative values indicating that reconstruction consistently reduced RMSE relative to the original simulated abnormal-gait trials.

The magnitude of RMSE reduction varied across joints and participant--task pairs. Hip movements and pelvis flexion/extension showed the largest effect sizes, consistent with the prominence of sagittal and transverse plane deviations across multiple instructed tasks (CD, HS, TF, TE, GG). Knee flexion/extension showed the smallest effect size ($r_{rb}{=}-0.41$), reflecting more variable correction patterns.

\begin{table}
\caption{Leave-one-participant-out sensitivity analysis. Each cell reports the rank-biserial correlation \(r_{rb}\), followed in parentheses by the percentage of the remaining participant--task pairs showing an RMSE reduction. Negative correlations indicate lower RMSE after reconstruction. The first row reports the original analysis of all 70 pairs; each subsequent row reports the analysis after excluding the seven pairs associated with the indicated participant. All leave-one-participant-out comparisons remained significant after Holm-Bonferroni correction (\(p_{\mathrm{Holm}}<0.05\)).}
\label{tab:loo_participant}

\begingroup
\setlength{\tabcolsep}{4pt}

\resizebox{\columnwidth}{!}{%
\begin{tabular}{@{}lcccc@{}}
\toprule
\textbf{Dataset} &
\textbf{Pelvis F/E} &
\textbf{Right hip A/A} &
\textbf{Right hip F/E} &
\textbf{Right knee F/E} \\
\midrule
Full analysis
& \(-0.96\) (95.7\%)
& \(-0.77\) (82.9\%)
& \(-0.80\) (84.3\%)
& \(-0.41\) (61.4\%) \\
\midrule
Without p001
& \(-0.95\) (95.2\%)
& \(-0.82\) (84.1\%)
& \(-0.81\) (84.1\%)
& \(-0.39\) (60.3\%) \\

Without p002
& \(-0.97\) (96.8\%)
& \(-0.79\) (84.1\%)
& \(-0.79\) (82.5\%)
& \(-0.39\) (58.7\%) \\

Without p003
& \(-0.96\) (95.2\%)
& \(-0.79\) (84.1\%)
& \(-0.79\) (84.1\%)
& \(-0.43\) (63.5\%) \\

Without p004
& \(-0.95\) (95.2\%)
& \(-0.78\) (84.1\%)
& \(-0.79\) (84.1\%)
& \(-0.43\) (61.9\%) \\

Without p005
& \(-0.95\) (95.2\%)
& \(-0.72\) (81.0\%)
& \(-0.83\) (85.7\%)
& \(-0.42\) (61.9\%) \\

Without p006
& \(-0.96\) (95.2\%)
& \(-0.74\) (81.0\%)
& \(-0.80\) (84.1\%)
& \(-0.43\) (61.9\%) \\

Without p007
& \(-0.96\) (95.2\%)
& \(-0.76\) (82.5\%)
& \(-0.78\) (82.5\%)
& \(-0.33\) (58.7\%) \\

Without p008
& \(-0.96\) (95.2\%)
& \(-0.74\) (81.0\%)
& \(-0.83\) (87.3\%)
& \(-0.38\) (60.3\%) \\

Without p009
& \(-0.96\) (96.8\%)
& \(-0.74\) (82.5\%)
& \(-0.77\) (82.5\%)
& \(-0.45\) (63.5\%) \\

Without p010
& \(-0.98\) (96.8\%)
& \(-0.77\) (84.1\%)
& \(-0.83\) (85.7\%)
& \(-0.43\) (63.5\%) \\
\bottomrule
\end{tabular}%
}

\endgroup
\end{table}

The leave-one-participant-out sensitivity analysis is reported in Table \ref{tab:loo_participant}. All four correction effects remained statistically significant after every participant exclusion following Holm-Bonferroni correction, with the largest adjusted \(p\)-value equal to \(0.023\). Across the ten exclusions, rank-biserial correlations ranged from \(-0.981\) to \(-0.951\) for pelvis flexion/extension, from \(-0.817\) to \(-0.723\) for right hip abduction/adduction, from \(-0.833\) to \(-0.771\) for right hip flexion/extension, and from \(-0.449\) to \(-0.330\) for right knee flexion/extension. The corresponding percentages of participant--task pairs showing an RMSE reduction ranged from \(95.2\%\) to \(96.8\%\), \(81.0\%\) to \(84.1\%\), \(82.5\%\) to \(87.3\%\), and \(58.7\%\) to \(63.5\%\), respectively. Thus, none of the aggregate correction effects depended on the inclusion of a single participant.

Importantly, reconstructed RMSE values remained above zero (non-zero deviation from the normative population mean trajectory), indicating that the model does not collapse all trials toward the training mean but instead aims to produce biomechanically plausible joint kinematics.

\section{Discussion}
\label{sec:discussion}
The framework exploits the biomechanical redundancy of the gait, as anatomical limits, dynamic coupling, and coordination patterns highly constrain joint configurations in humans during such activity.
During inference, \textit{Pass 1} systematically tests each joint's contribution by removing it from the evidence. Joints whose absence significantly alters the predicted kinematics reveal themselves as inconsistent with the learned biomechanical prior. Prior anomaly detection approaches \cite{nguyen2016skeleton, duan2024fsgait} define abnormality as elevated reconstruction error between the input pose and the model's output, and a joint is flagged when the model output diverges from the input. This measure does not distinguish biomechanical inconsistency from pose estimation noise and individual kinematic variability, since a high reconstruction error may simply reflect an unusual but biomechanically valid configuration that the model has not seen, or a reconstruction/input artefact unrelated to any true kinematic deviation. The proposed badness score instead uses a different operational definition of joint inconsistency, quantifying how much a joint's observed configuration contradicts expectations built from the remaining body context and temporal dynamics, asking not whether the joint can be reproduced, but whether it is coherent with the normative learnt prior and the spatiotemporal constraints imposed by the remaining skeleton.\\
\textit{Pass 2} then leverages these findings to perform targeted correction, asking the model to reconstruct flagged joints from the reliable biomechanical context. This is why the approach works without disease labels; abnormality is defined as a joint configuration that cannot be explained by normative spatiotemporal constraints, rather than by matching predefined anomaly templates.
This definition of abnormality requires the model to learn a distributional manifold of biomechanically plausible configurations shaped by the range of joint coordination patterns, angular velocities, and inter-joint dependencies observed across diverse normative individuals. During reconstruction, the model does not regress toward the population mean but instead generates configurations that lie within the learned biomechanically plausible manifold, constrained by the local spatiotemporal context. \\
The experimental validation was designed to test the framework's robustness across diverse biomechanical deviations. Results were presented at two complementary levels to assess normative preservation and deviation correction.
Representative per-task examples, Figure \ref{fig:skel}, Figure \ref{fig:grid}, and Table \ref{tab:rmse}, were selected based on visual inspection of the execution quality to identify participants whose movements most closely matched the instructions provided (Section \ref{sec:experiments}). This selection, performed independently of model performance to avoid cherry-picking, provides interpretable examples of simulated movements that clearly reflected the respective instructions. The normative panel was then randomly selected from the remaining participants not assigned to any simulated abnormal-gait example, ensuring an unbiased qualitative reference.

Statistical analysis on the simulated abnormal-gait trials, shown in Figure \ref{fig:non-n}, then aggregated all 70 participant-task pairs regardless of execution fidelity. Even though visual inspection revealed substantial variability in how participants performed the instructed tasks, all trials were retained. Variable execution quality, including incomplete, exaggerated, or atypical realizations, provides a more realistic test of generalization than perfectly controlled, homogeneous impairments would. Real patient populations exhibit similar heterogeneity due to disease stage, compensatory strategies, comorbidities, and individual motor control capacity, underscoring the relevance of test dataset variability. The model's design objective, indeed, is anomaly detection and correction of any deviation from learned normative structure, not classification of predefined anomaly types.\\
Importantly, the aggregate results do not imply uniform correction performance across the seven instructed tasks and cannot establish which simulated gait disorders benefit most from correction. A comparison among the task-defined groups could not determine whether the observed differences arose from correction performance or from differences in the joints recruited and in the magnitude, timing, and compensatory structure of the movements produced by each participant. A biomechanically meaningful subgroup analysis would require independent trial-level annotations of these characteristics, which were not collected in the present protocol.\\
The representative normative trial showed no flagged joints and near-perfect overlap between original and reconstructed skeletons (Figure~\ref{fig:skel}), with original and reconstructed trajectories tracking each other closely and remaining within the normative reference band throughout the gait cycle (Figure~\ref{fig:grid}), providing a qualitative example of normative preservation. Across all ten held-out normative trials, the operational false-selection rate was \(1/60\), or \(1.7\%\). The single threshold exceedance involved the left knee of participant~7 and was traced to a knee-angle singularity associated with inaccurate toe-point detection. Although operationally counted as a false selection relative to the participant's normative gait, it reflected an actual inconsistency in the input kinematics rather than an arbitrary score fluctuation, showing that Pass~1 responds to deviations from the learned normative prior regardless of whether they originate from the movement itself or from upstream pose-estimation errors.\\
Quantitatively, the RMSE differences computed between the mean of original and reconstructed normative trials remained below $0.82\degree$ across all joints (Table~\ref{tab:rmse}), and bootstrap equivalence testing confirmed statistical equivalence within $\delta{=}1.5\degree$ for all four joint angles after Holm-Bonferroni correction (Figure~\ref{fig:n}).
In the representative simulated abnormal-gait examples, the selected joints were biomechanically compatible with the dominant visible deviations (Figure~\ref{fig:skel}). The present simulated-gait protocol did not provide independent joint-level ground truth identifying which joints were actually affected in each trial, preventing quantitative validation of Pass~1 localization accuracy. Reconstruction visibly reduced joint angle deviations from the normative band while retaining the temporal organization visible in the input trajectories (Figure~\ref{fig:grid}). Wilcoxon signed-rank tests confirmed that these reductions were statistically significant across all four analyzed joints ($p < 0.003$, large effect sizes; Figure~\ref{fig:non-n}). 
The leave-one-participant-out analysis further showed that all four effects remained significant regardless of which participant was excluded, with only limited variation in the effect sizes and percentages of improved participant--task pairs. This indicates that the aggregate findings were not determined by any individual participant.\\
Taken together, these results directly reflect how the model targets plausibility rather than conformity to a specific reference trajectory. Normative trials remain largely unchanged because they already occupy regions of the learned manifold, and the model does not introduce artificial deviations when processing biomechanically valid movement patterns, while pathological patterns shift toward the manifold boundary without collapsing to a single "average" normative gait.\\
Correction efficacy, however, varied systematically across joints. The 7-frame window architecture limits access to the full gait cycle (40-60 frames, 1.3-2 seconds), as it covers approximately 15--20\% of one cycle, and therefore restricts its ability to learn coordination patterns spanning an entire gait cycle. It also cannot represent cycle-to-cycle consistency, which would require context extending across multiple complete cycles. Without full-cycle context, the model may have limited ability to distinguish exaggerated compensation from normal speed-related variation. When other local cues suggest biomechanical validity, the model may preserve or even amplify certain deviations, particularly in distal joints like the knee. The restricted temporal context may therefore have contributed to the smaller and more variable correction effects observed for knee flexion/extension compared with the proximal joints. However, its contribution was not isolated experimentally, and knee-angle singularities in the training data caused by inaccurate toe-point localization provide an additional technical explanation. Future work should directly evaluate the effect of temporal context through controlled comparisons with phase-normalized full-cycle attention and hierarchical multi-scale architectures combining local-window representations with cycle-level context.\\
Beyond these architectural considerations, additional specifications warrant discussion. First, the abnormal-gait patterns were simulated by normative participants following verbal instructions and a brief familiarization exercise, not genuine patients with neurological or orthopedic impairments. While this approach introduced realistic motor variability and allowed controlled testing of specific instructed deviations, the kinematic signatures may differ quantitatively from those of clinical populations. The present results therefore represent a proof of concept on simulated biomechanical deviations and cannot establish performance on clinical gait shaped by chronic neuromuscular adaptations, disease progression, comorbidities, and long-term compensatory strategies. Validation on patient cohorts is required before drawing conclusions about clinical performance. Second, the normative reference was derived from 150 adults without gait impairments walking at self-selected speeds on level ground. The model was designed to represent normative gait in adults, and pediatric and older-adult gait, which require age-specific normative characterization, were outside the intended population of this study. Nevertheless, the overall cohort was concentrated around early adulthood (\(31.1\pm6.2\) years, range 22--57), and future datasets should include more participants near the younger and older ends of the intended adult population. BMI and ethnicity were not used as exclusion criteria but were not recorded; consequently, the composition of these subgroups and generalization across them cannot be quantified. Third, markerless pose estimation accuracy remains a limiting factor. Although the choice of a markerless system reflected a deliberate trade-off between acquisition scalability and detection accuracy, the Real-Move system exhibited reduced reliability for distal keypoints, leading to the exclusion of ankles from the anomaly detection process and producing kinematic singularities in knee-angle trajectories due to inaccurate toe-point localization. The constrained interpolation applied during preprocessing addresses temporary missing samples, while the two-pass masking strategy improves robustness to missing and corrupted data. However, neither mechanism can recover reliable ankle kinematics from systematic pose estimation errors in toe-point joints, which may propagate through the biomechanical chain, affecting reconstruction quality. Consequently, the current framework cannot directly detect or correct ankle-level abnormalities such as reduced dorsiflexion or foot drop, and extending the framework to ankle-level analysis is therefore an important direction for future work.

\section{Conclusions}
\label{sec:conclusion}
This work presented a transformer-based framework for anomaly detection and correction in gait kinematics that operates without disease-labeled training data. By exploiting biomechanical redundancy, the two-pass procedure identifies joints inconsistent with learned normative spatiotemporal constraints and reconstructs them from reliable context, generating a normative twin of the input that reflects how the observed gait would appear if the detected deviations were removed. Proof-of-concept evaluation on simulated abnormal-gait tasks showed statistically significant overall reductions in angular deviation across the analyzed joints while preserving unseen normative kinematics. The leave-one-participant-out sensitivity analysis further showed that all four correction effects remained significant after every participant exclusion, indicating that no individual participant determined the aggregate findings.

The present methodological contribution consists of joint-level identification of deviations from learned normative gait patterns and subject-conditioned normative reconstruction rather than diagnostic classification.

Accordingly, the framework should be regarded as a methodological basis for future clinical investigation. Its relevance to rehabilitation decisions and repeated patient assessment must be established using genuine patient gait, a more representative normative reference, and prospective evaluation within clinical practice. Further technical development should extend the temporal context to the full gait cycle and improve distal-keypoint measurement sufficiently to evaluate ankle-level outputs.

\printcredits

\bibliographystyle{cas-model2-names}

\bibliography{cas-refs.bib}

\end{document}